\documentclass{article} % For LaTeX2e
\usepackage{iclr2025_conference,times}

\def\eqref#1{equation~\ref{#1}}
\def\1{\bm{1}}

\usepackage{hyperref}       % hyperlinks
\usepackage{url}            % simple URL typesetting
\usepackage{booktabs}       % professional-quality tables
\usepackage{amsfonts}       % blackboard math symbols
\usepackage{nicefrac}       % compact symbols for 1/2, etc.
\usepackage{microtype}      % microtypography
\usepackage{xcolor}         % colors
\usepackage{graphicx} 
\usepackage{subcaption}
\usepackage{multirow}
\usepackage{algorithm}
\usepackage{algorithmic}
\usepackage{bbm}
\usepackage{arydshln}
\usepackage{booktabs}
\usepackage{caption}
\usepackage{bm}
\usepackage{arydshln}
\usepackage{stfloats}
\usepackage{graphicx}
\usepackage{subcaption}
\usepackage{natbib}
\usepackage{algorithm}
\usepackage{algorithmic}
\usepackage{bbm}
\usepackage{arydshln}
\usepackage{amsmath} 
\usepackage{booktabs}
\usepackage{colortbl}
\usepackage{hyperref}
\usepackage{url}
\usepackage{graphicx}
\usepackage{colortbl}
\usepackage{multicol}
\usepackage{wrapfig}

\hypersetup{
    colorlinks=false,
    linkcolor=blue,
    citecolor=blue,
    urlcolor=blue,
    pdfborder={0 0 0}  % 去掉框框
}

\title{IMDPrompter: Adapting SAM to Image Manipulation Detection by Cross-View Automated Prompt Learning}

\author{Quan Zhang$^{1}$\thanks{These authors contributed equally.}, Yuxin Qi$^{2}$\footnotemark[1], Xi Tang$^{1}$, Jinwei Fang$^{3}$, Xi Lin$^{2}$, Ke Zhang$^{1}$\thanks{Corresponding authors.}, Chun Yuan$^{1}$\footnotemark[2] \\
$^1$Tsinghua University\,\,
$^2$Shanghai Jiao Tong University \\
$^3$University of Science and Technology of China\\
\texttt{\{zhangqua22, ke-zhang19\}@mails.tsinghua.edu.cn,}\\
\texttt{qiyuxin98@sjtu.edu.cn, yuanc@sz.tsinghua.edu.cn} 
}

\iclrfinalcopy % Uncomment for camera-ready version, but NOT for submission.
\begin{document}

\maketitle

\begin{abstract}
Using extensive training data from SA-1B, the Segment Anything Model (SAM) has demonstrated exceptional generalization and zero-shot capabilities, attracting widespread attention in areas such as medical image segmentation and remote sensing image segmentation. However, its performance in the field of image manipulation detection remains largely unexplored and unconfirmed. There are two main challenges in applying SAM to image manipulation detection: a) reliance on manual prompts, and b) the difficulty of single-view information in supporting cross-dataset generalization. To address these challenges, we develops a cross-view prompt learning paradigm called IMDPrompter based on SAM. Benefiting from the design of automated prompts, IMDPrompter no longer relies on manual guidance, enabling automated detection and localization. Additionally, we propose components such as Cross-view Feature Perception, Optimal Prompt Selection, and Cross-View Prompt Consistency, which facilitate cross-view perceptual learning and guide SAM to generate accurate masks. Extensive experimental results from five datasets (CASIA, Columbia, Coverage, IMD2020, and NIST16) validate the effectiveness of our proposed method.
\end{abstract}

\section{Introduction}

With the continuous emergence of powerful editing tools \cite{zhang2025imdprompter}, image manipulation has become unprecedentedly simple. These new opportunities have sparked the creativity of both benevolent and malicious users. In the past, organizing multimedia misinformation activities required complex skills, with attackers limited to splicing, copying, or deleting objects.Recently, deep learning has advanced at a rapid pace 
, leading to the development of more user-friendly and powerful image editing tools. As a result, users can now swiftly generate images of fictional characters or create highly convincing deepfakes. Generative models can produce realistic image edits based on natural language prompts, perfectly matching inserted elements with the style and lighting of the environment \cite{avrahami2023blended,nichol2021glide}.

However, the risks of these tools falling into the wrong hands are evident. In recent years, governments and funding agencies have shown increasing interest in developing forensic tools capable of addressing such attacks, particularly focusing on local image tampering that alters the semantic content of images \cite{le2021openforensics}. In response to these challenges, the fields of multimedia forensics and related sciences have rapidly expanded, proposing various methods and tools for image manipulation detection (IMD) and localization \cite{guillaro2023trufor}. Despite significant progress in this area, the performance of state-of-the-art detectors in practical applications remains insufficient, primarily limited by several shortcomings that require further research: a) limited generalization ability; b) limited robustness; c) poor detection performance.

To address the shortcomings of existing IMD methods, we turn our attention to foundational models \cite{kirillov2023segment, radford2021learning, jia2021scaling, sharif2014cnn}. Thanks to large-scale pretraining, 
foundational models like GPT-4 
\cite{achiam2023gpt}, Flamingo 
 \cite{alayrac2022flamingo}, and SAM \cite{kirillov2023segment} have made significant strides and contributed importantly to societal advancements. Among them, SAM, trained on one billion masks, demonstrates exceptional 
generalization ability, inspiring our research. However, using SAM for iimage manipulation detection tasks presents significant challenges: a) reliance on manual prompts: SAM's interactive framework requires predefined prompts for input images, such as points, boxes, or masks. As a category-agnostic segmentation method, these limitations hinder SAM's ability to achieve fully automated understanding in IMD tasks; b) single-view information struggles to support cross-dataset generalization: previous studies \cite{zhou2020generate} have reported that DeepLabv2 \cite{chen2017deeplab} trained on the CASIAv2 \cite{dong2013casia} dataset performed well on the CAISAv1 \cite{dong2013casia} dataset (sourced from CASIAv2) but poorly on the non-sourced COVER \cite{wen2016coverage} dataset. Similar behavior has also been observed with FCN \cite{long2015fully} in this research.

% Please add the following required packages to your document preamble:
% \usepackage{multirow}

\begin{wrapfigure}{r}{0.55\textwidth}
    \centering
    \includegraphics[width=0.55\textwidth]{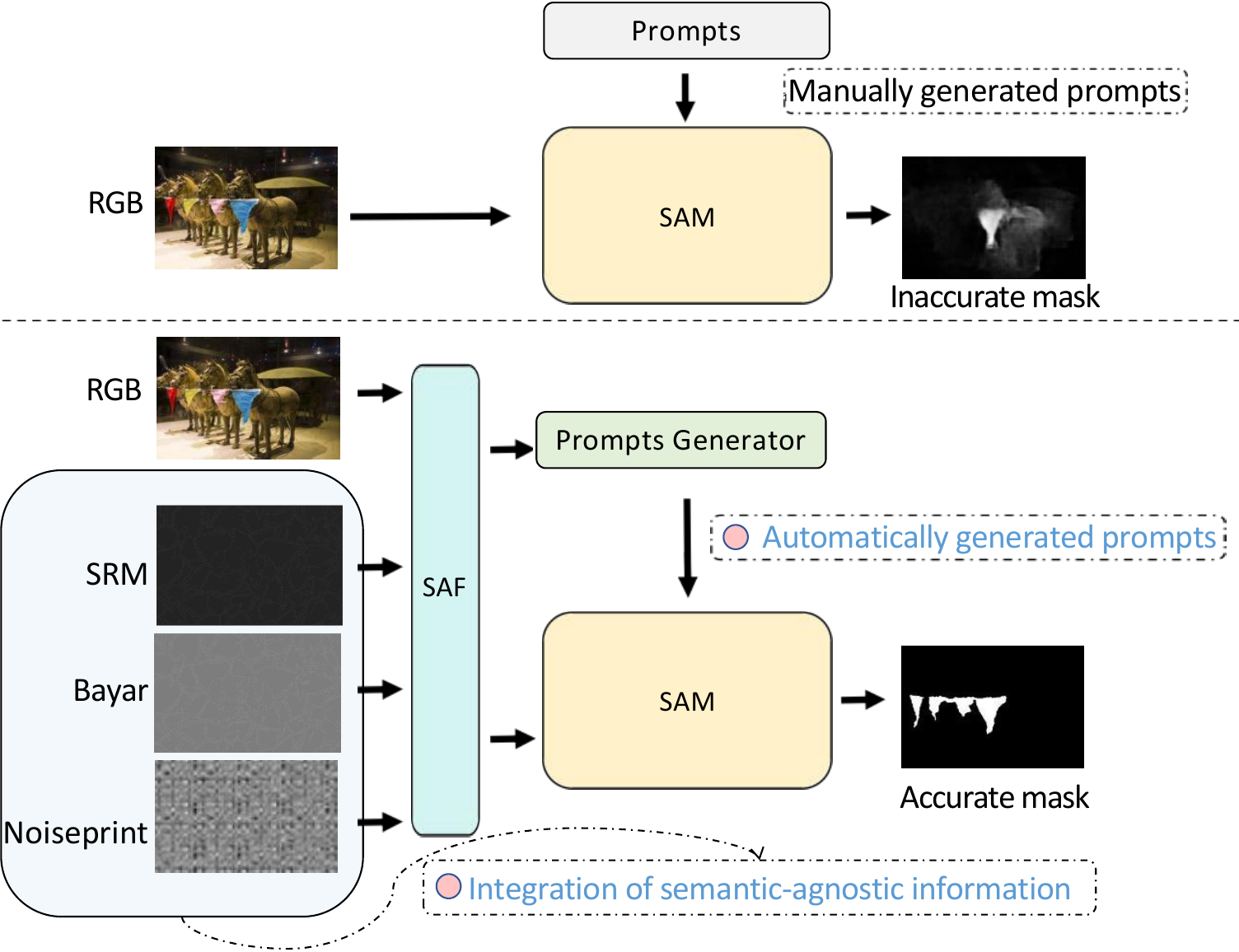}
    \caption{Improvements of proposed IMDPrompter over existing pipelines of baselines: (a). Automated prompt learning without the need for manual input. (b). Flexible integration of semantic-agnostic information crucial for Image Manipulation Detection.}
    \label{fig:intro}
\end{wrapfigure}

To activate SAM's IMD capabilities, we propose a cross-view automated prompt learning paradigm called IMDPrompter, aimed at learning how to generate prompts to enhance SAM's functionality: a) Inspired by \cite{battle2024unreasonable} using automated prompt optimizers to improve the performance of LLMs, we constructed the automated prompt learning paradigm IMDPrompter, which generates prompt information automatically; b) To enhance SAM's cross-dataset generalization ability, we introduced three noise views: SRM filtering, Bayer Conv, and Noiseprint, aiming to generate enhanced prompt information from multiple perspectives. 
% To integrate information from multiple views, previous methods \cite{zhai2023towards}  opted for weighted averaging of outputs from multiple views. 
To integrate information from multiple views, previous methods \cite{zhai2023towards} employed a strategy of weighted averaging of the outputs from multiple views.
However, due to the inaccuracy of prompts generated from individual views, simple weighted averaging can smooth out the accurate prompts from the optimal view to suboptimal ones. To preserve the accurate prompts from the optimal view, we propose an Optimal Prompt Selection (OPS) module based on minimizing localization loss; meanwhile, we noted that ideally, the prompt information from each view should converge toward the optimal prompt. Based on this motivation, we constructed a Cross-View Prompt Consistency (CPC) constraint to achieve alignment enhancement across views. For the mask generation process, we developed an Attention-based Cross-view Feature Perception (CFP) module and a Multi-Layer Perceptron-based Prompt Mixing  module (PMM) to achieve the fusion of cross-view information and the integration of multiple types of prompts sequentially.

In summary, our main contributions are as follows.
\begin{itemize}
\item We are the first to apply SAM to the field of image manipulation detection and propose an automated prompt learning paradigm, IMDPrompter, eliminating the original SAM's reliance on manual prompts.
\item We propose modules such as Optimal Prompt Selection and Cross-View Prompt Consistency Constraint, achieving alignment enhancement across views.
\item We propose Cross-view Feature Perception and Prompt Mixing modules, achieving the fusion of cross-view information and the integration of multiple types of prompts.
\item We demonstrate extensive results across five different image manipulation detection datasets, thoroughly validating the strong in-distribution and out-of-distribution image manipulation detection and localization capabilities of IMDPrompter.
\end{itemize}

\section{Related Work}
\noindent\textbf{Image Manipulation Detection.} Currently, methods for image manipulation detection can be broadly categorized into two types, primarily distinguished by their recognition of manipulated artifacts. Some techniques \cite{wu2019mantra, chen2021image, wu2022robust, bi2019rru, hu2020span, yang2020constrained, marra2020full} rely on detecting abnormal features and often use high-pass noise filters \cite{yang2020constrained, li2019localization} to suppress content information. Other methods \cite{park2018double, kwon2022learning, mareen2022comprint} attempt to detect inconsistencies in compression within tampered images, as they assume different compression Quality Factors (QFs) before and after the operation. Additionally, some researchers focus their attention on camera-based artifacts, such as model fingerprints \cite{mareen2022comprint, cozzolino2019noiseprint, cozzolino2015splicebuster}.

\noindent\textbf{Foundational Models.} In recent years, foundational models have sparked a tremendous transformation in the field of artificial intelligence . These models, trained on extensive datasets, have demonstrated impressive generalization capabilities across various scenarios \cite{kirillov2023segment,cai2024biosam,liu2024pq,pan2024conv,pan2024codev,pan2025code}. Renowned models such as Chat-GPT \cite{ouyang2022training}, GPT-4 \cite{achiam2023gpt}, and Stable Diffusion \cite{rombach2022high} have further propelled the development of artificial intelligence, making significant contributions to human civilization and exerting considerable influence across various industries. 
Inspired by the success of foundational models in natural language processing (NLP), researchers have begun exploring their potential applications in computer vision. While most of these models are aimed at extracting accessible knowledge from freely available data \cite{alayrac2022flamingo, radford2021learning, chen2023ovarnet}, the recent SAM model \cite{kirillov2023segment} adopts an innovative approach by constructing a data engine where the model co-develops annotations with environmental datasets. SAM uniquely leverages a vast collection of masks, showcasing robust generalization capabilities. However, it was initially designed as a task-agnostic segmentation model, requiring prompts (i.e., inputs of prior points, bounding boxes, or masks), and therefore does not directly facilitate end-to-end automated segmentation perception.
This paper does not delve into the design and training of foundational image manipulation detection models; instead, we explore the potential of utilizing SAM's powerful universal segmentation capabilities for image manipulation detection and localization. Furthermore, the proposed method of learning prompts can be extended to other visual foundational models beyond SAM.

\noindent\textbf{Prompt Learning.} In the past, machine learning tasks were primarily focused on fully supervised learning, where task-specific models were trained only on labeled instances of the target task  \cite{krizhevsky2017imagenet}. However, over time, there has been a significant shift in learning paradigms, transitioning from fully supervised learning towards \textit{pretraining and fine-tuning }approaches for downstream tasks. This shift allows models to leverage general features acquired during pretraining \cite{russakovsky2015imagenet, simonyan2014very, he2016deep}.
More recently, with the advent of foundational models, a new paradigm has emerged known as \textit{pretraining and prompting} \cite{chen2023ovarnet, lester2021power, liu2023pre, zhou2022learning}. In this paradigm, researchers no longer train models specifically for downstream tasks but instead redesign inputs using prompts to reformulate the downstream tasks to align with the original pretraining task \cite{radford2021learning, devlin2018bert, radford2019language}. Prompting helps to reduce semantic gaps, bridge the gap between pretraining and fine-tuning, and prevent overfitting of the heads. Since the advent of GPT-3 \cite{brown2020language}, prompting has evolved from traditional discrete \cite{liu2023pre} and continuous prompt constructions \cite{chen2023ovarnet, zhou2022learning} to large-scale model-centric contextual learning \cite{alayrac2022flamingo}, instruction tuning \cite{liu2024visual, peng2023instruction, gupta2022instructdial}, and chaining approaches \cite{wei2022chain, wang2022self, zhang2022automatic}.
Currently, methods for constructing prompts include manual templates, heuristic-based templates, generation, fine-tuning word embeddings, and pseudo-labeling \cite{liu2023pre, wang2022learning}. In this paper, we propose a prompt generator for generating prompts compatible with SAM.
% \label{headings}

% First level headings are in small caps,
% flush left and in point size 12. One line space before the first level
% heading and 1/2~line space after the first level heading.

% \subsection{Headings: second level}

% Second level headings are in small caps,
% flush left and in point size 10. One line space before the second level
% heading and 1/2~line space after the second level heading.

% \subsubsection{Headings: third level}

% Third level headings are in small caps,
% flush left and in point size 10. One line space before the third level
% heading and 1/2~line space after the third level heading.

\section{Proposed Method}
\begin{figure*}[ht]
	\centering
	\includegraphics[width=0.95\linewidth]{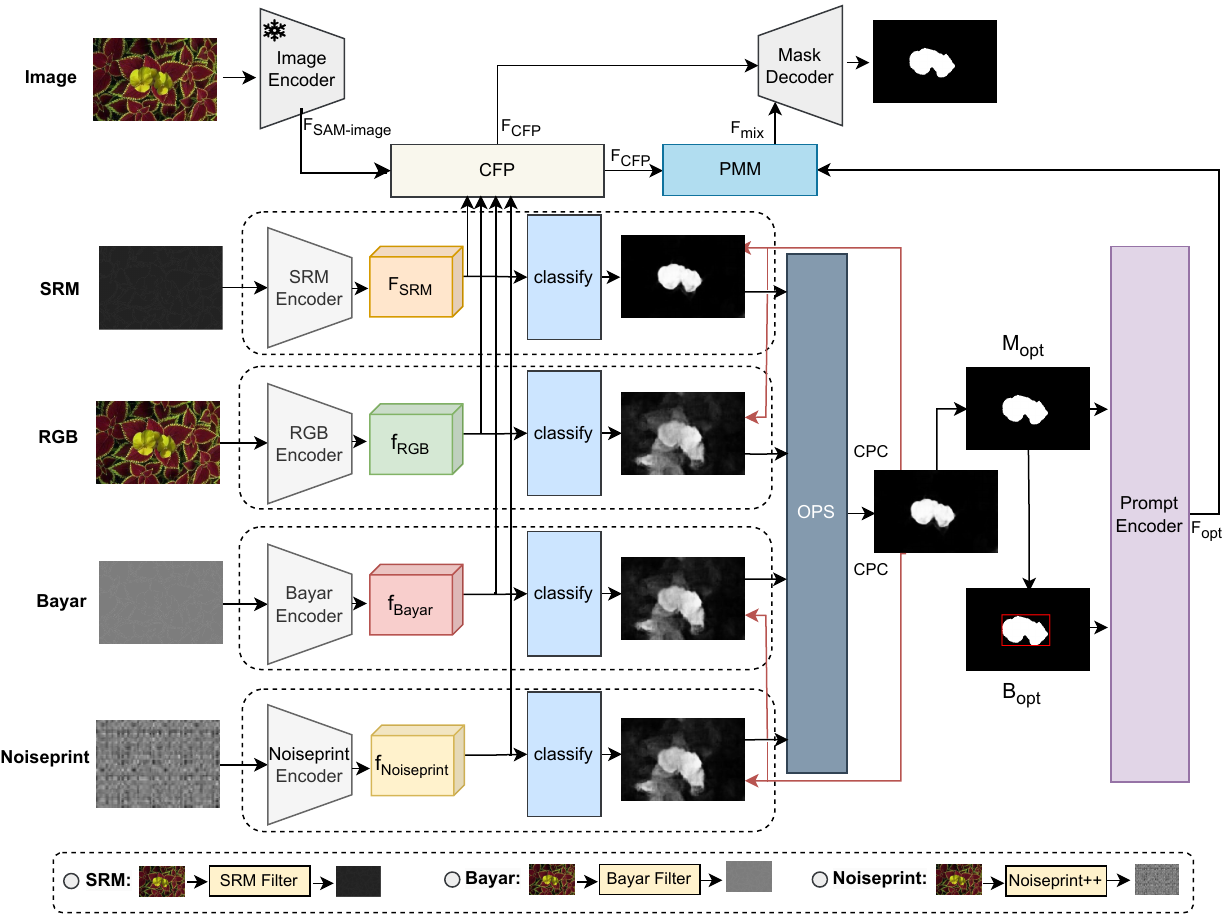}
	\caption{Overall framework of IMDPrompter. The prompter part consists of four views: RGB, SRM, Bayar and Noiseprint. OPS selects the optimal prediction from the four views to generate the best prompt. CPC enhances cross-view consistency. CFP achieves cross-view feature perception fusion. PMM achieves a mixture of multiple prompt information.}
	\label{fig:pipe}
\end{figure*}

% \begin{figure}[t]
% 	\centering
% 	\includegraphics[width=0.7\linewidth]{sec/fig/sae.pdf}
% 	\caption{Architecture of the Cross-view Feature Perception (CFP) unit.}
% 	\label{fig:CFP}
% \end{figure}

As illustrated in Figure \ref{fig:pipe}, we introduce IMDPrompter, an automated prompt learning framework leveraging cross-view perception to unlock SAM's detection and localization capabilities. It employs an optimal prompt selection module to identify the best prompts, enhances alignment via cross-view prompt consistency, and achieves efficient image manipulation detection and localization through feature fusion and prompt mixing.

\subsection{Multi-view feature representation}

Current technology \cite{zhou2020generate} shows that while RGB views work well for in-domain (IND) manipulation detection, they perform poorly in out-of-domain (OOD) detection. Moreover, using noise views to capture semantic-agnostic information enhances performance. Therefore, relying solely on RGB view data is inadequate for detecting and localizing manipulations. To address this, we incorporate three semantic-agnostic views: SRM, Bayer, and Noiseprint noise views, as outlined below:

\begin{equation}
F_{\text{sam}} = \Phi_{\text{sam-img}}(\mathcal{I}),
\end{equation}
\begin{equation}
f_{\text{RGB}} = \Phi_{\text{Seg-RGB}}(\mathcal{I}),  f_{\text{SRM}} = \Phi_{\text{Seg-SRM}}\left(\Phi_{\text{SRM}}(\mathcal{I})\right),  f_{\text{Bayer}} = \Phi_{\text{Seg-Bayer}}\left(\Phi_{\text{Bayer}}(\mathcal{I})\right),
\end{equation}
\begin{equation}
f_{\text{Noiseprint}} = \Phi_{\text{Seg-Noiseprint}}\left(\Phi_{\text{Noiseprint}}(\mathcal{I})\right).
\end{equation}
In the system, $\mathcal{I}$ represents the input image, and $\Phi_{\text{sam-img}}$ denotes the SAM image encoder. $F_{\text{sam}}$ represents the features encoded by the SAM image encoder. $\Phi_{\mathrm{SRM}}$, $\Phi_{\text{Bayer}}$ and $\Phi_{\text{Noiseprint}}$ respectively represent the SRM noise map extractor, the Bayer noise map extractor and the Noiseprint noise map extractor. $\Phi_{\text{Seg-RGB}}$, $\Phi_{\text{Seg-SRM}}$, $\Phi_{\text{Seg-Bayer}}$ and $\Phi_{\text{Seg-Noiseprint}}$ each represent the segmenter for the RGB view, SRM view, Bayer view and Noiseprint view, with all four segmenters having identical structures but non-shared parameters. $f_{\mathrm{RGB}}$, $f_{\mathrm{SRM}}$, $f_{\mathrm{Bayer}}$ and $f_{\mathrm{Noiseprint}}$ represent the features from the RGB, SRM, Bayer, and Noiseprint views, respectively. 

\subsection{Optimal Prompt Selection}
% In the process of generating masked prompts, we utilize the mask probability distributions from four views, $P_{\mathrm{RGB}}$, $P_{\mathrm{SRM}}$, $P_{\mathrm{Bayar}}$, and $P_{\mathrm{Noiseprint}}$, and construct an integrated mask probability distribution $P_{\text{Ens}}$. We then select the optimal mask probability segmentation $P_{\mathrm{opt}}$ based on the principle of minimizing segmentation loss. Further, we obtain the mask $M_{\mathrm{opt}}$ and bounding box prompts $\mathcal{B}_{\mathrm{opt}}$, which are input into SAM's prompt encoder to obtain the prompt encoding $F_{\mathrm{opt}}$. The process is as follows:
In the masked prompt generation process, we integrate mask probability distributions from four views: $P_{\mathrm{RGB}}$, $P_{\mathrm{SRM}}$, $P_{\mathrm{Bayar}}$, and $P_{\mathrm{Noiseprint}}$, to form an ensemble distribution $P_{\text{Ens}}$. We select the optimal mask probability segmentation $P_{\mathrm{opt}}$ by minimizing segmentation loss, and derive the corresponding mask $M_{\mathrm{opt}}$ and bounding box prompts $\mathcal{B}_{\mathrm{opt}}$. These are fed into SAM's prompt encoder to obtain the prompt encoding $F_{\mathrm{opt}}$. The process is outlined as follows:
\begin{equation}
P_{\mathrm{RGB}} = \Phi_{\mathrm{RGB}-\mathrm{CLS}}(f_{\mathrm{RGB}}),  P_{\mathrm{SRM}} = \Phi_{\mathrm{SRM-CLS}}(f_{\mathrm{SRM}}),  P_{\mathrm{Bayer}} = \Phi_{\mathrm{Bayer-CLS}}(f_{\mathrm{Bayer}}),
\end{equation}
\begin{equation}
P_{\mathrm{Noiseprint}} = \Phi_{\mathrm{Noiseprint-CLS}}(f_{\mathrm{Noiseprint}}),
\end{equation}
\begin{equation}
P_{\text{Ens}} = \frac{P_{\mathrm{RGB}} + P_{\mathrm{SRM}} + P_{\mathrm{Bayer}}+P_{\mathrm{Noiseprint}}}{4},
\end{equation}
\begin{equation}
P_{\text{opt}} = \underset {P \in \mathcal{P}}{\operatorname{argmin}} \mathcal{L}_{\mathrm{Seg}}(P, G),  \mathcal{P}=\{P_{\mathrm{RGB}}, P_{\mathrm{SRM}}, P_{\mathrm{Bayer}},P_{\mathrm{Noiseprint}}, P_{\text{Ens}}\},
\end{equation}
\begin{equation}
M_{\mathrm{opt}} = \Phi_{\mathrm{mask}}(P_{\mathrm{opt}}),  \mathcal{B}_{\mathrm{opt}} = \Phi_{\mathrm{box}}(M_{\mathrm{opt}}),
\end{equation}
\begin{equation}
F_{\mathrm{opt}} = \Phi_{\mathrm{p-enc}}(M_{\mathrm{opt}}, \mathcal{B}_{\mathrm{opt}}),
\end{equation}
where $\Phi_{\mathrm{RGB-CLS}}$, $\Phi_{\mathrm{SRM-CLS}}$, $\Phi_{\mathrm{Bayer-CLS}}$ and $\Phi_{\mathrm{Noiseprint-CLS}}$ represent classifiers for the RGB, SRM, Bayer, and Noiseprint views, respectively, $G$ represents the one-hot encoded mask labels, $\mathcal{L}_{\mathrm{Seg}}$ is the segmentation loss function, $\Phi_{\mathrm{mask}}$ represents the mask generation operation, $\Phi_{\mathrm{box}}$ is the bounding box generation operation, and $\Phi_{\mathrm{p-enc}}$ represents SAM's prompt encoder.

\subsection{Cross-View Prompt Consistency}
Ideally, the segmentation masks from the four views should be consistent with the optimal segmentation mask. Therefore, we constructed a cross-view prompt consistency enhancement loss to achieve enhanced prompt consistency across views. The $\mathrm{CPC}$ loss function is expressed as follows:
\begin{equation}
\mathcal{L}_{\mathrm{CPC}} = \mathcal{L}_{\mathrm{Seg}}(P_{\mathrm{RGB}}, P_{\text{opt}}) + \mathcal{L}_{\mathrm{Seg}}(P_{\mathrm{SRM}}, P_{\text{opt}}) + \mathcal{L}_{\mathrm{Seg}}(P_{\mathrm{Bayer}}, P_{\text{opt}}) +
\mathcal{L}_{\mathrm{Seg}}(P_{\mathrm{Noiseprint}}, P_{\text{opt}}),
\end{equation}
where $\mathcal{L}_{\mathrm{Seg}}$ is the segmentation loss function.

\subsection{Cross-view Feature Perception}
% \vspace{-5pt}
\begin{wrapfigure}{r}{0.48\textwidth}
    \centering
    \includegraphics[width=0.48\textwidth]{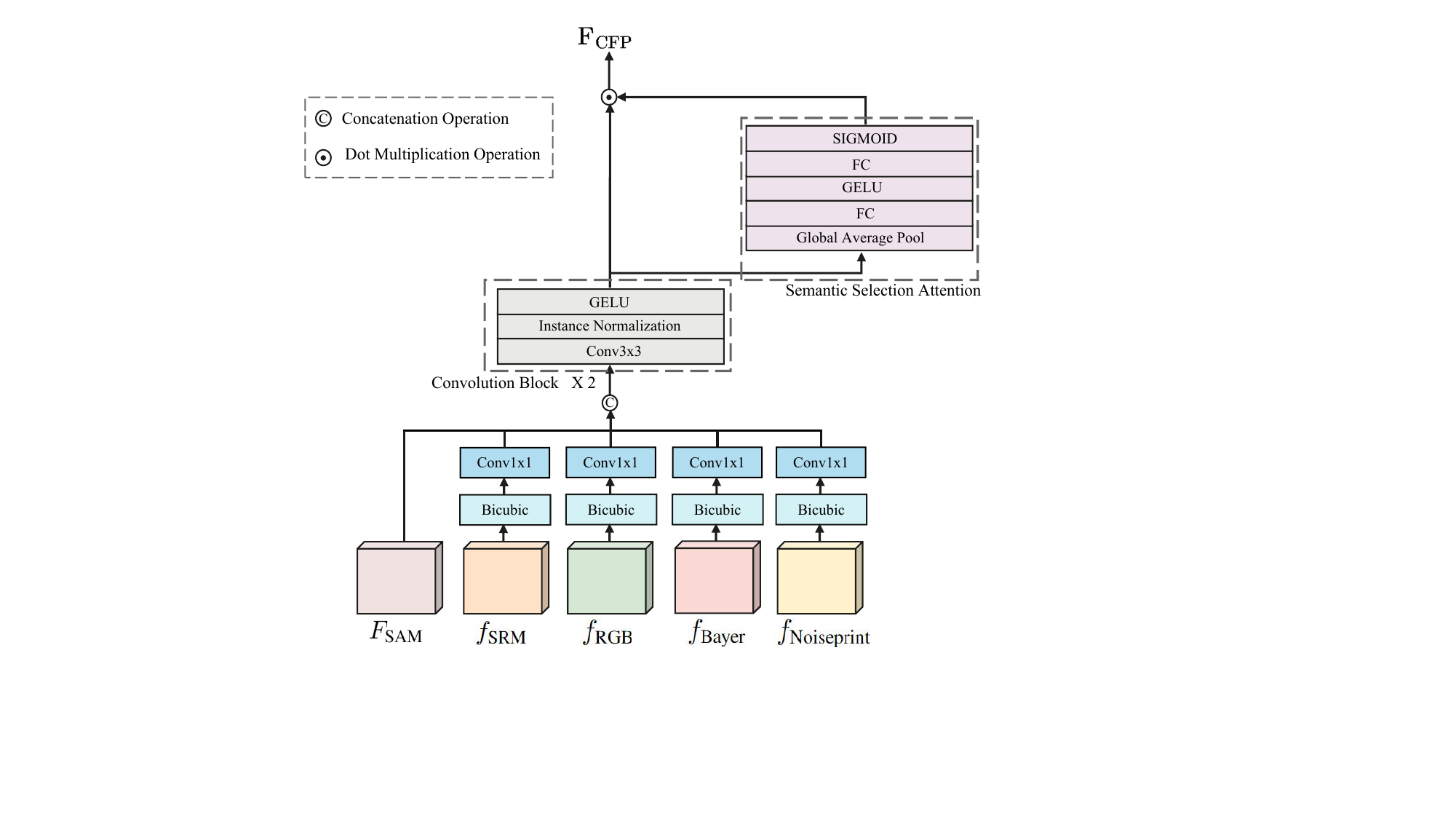}
    \caption{Architecture of the Cross-view Feature Perception (CFP) unit.}
    \label{fig:CFP}
\end{wrapfigure}

% Regarding the decoding process of SAM, relying solely on RGB view information is insufficient for generating accurate segmentation masks. Therefore, we propose a cross-view feature perception (CFP) module that integrates cross-view perception features from RGB, SRM, Bayer, and Noiseprint views, providing rich feature representations for the decoding process of SAM. The entire process is illustrated in Figure \ref{fig:CFP}. $\Phi_{\mathrm{CFP}}$ represents the cross-view feature perception module, while $F_{\mathrm{CFP}}$ denotes the features after cross-view perception fusion. The structure diagram of the CFP module is shown in Figure \ref{fig:CFP}.

% First, the features $f_{\text{RGB}}$, $f_{\text{SRM}}$, $f_{\text{Bayer}}$, and $f_{\text{Noiseprint}}$ are aligned with $F_{\text{sam}}$ through bicubic interpolation, followed by channel concatenation with $F_{\text{sam}}$ after passing through a $1 \times 1$ convolution. The concatenated features then pass through two stacked convolution blocks (comprising $3 \times 3$ convolution, instance normalization, and GELU activation), and the output from the semantic selection attention branch (consisting of global average pooling, a fully connected layer, GELU, and sigmoid) is weighted onto the shortcut branch to produce the integrated feature $F_{\text{CFP}}$. Through the CFP module, features from multiple views are effectively integrated.

To enhance the segmentation accuracy of SAM, we propose a Cross-View Feature Perception (CFP) module that integrates features from RGB, SRM, Bayer, and Noiseprint views. This module enriches the feature representations for SAM's decoding process, as shown in Figure \ref{fig:CFP}. Here, $\Phi_{\mathrm{CFP}}$ denotes the CFP module, and $F_{\mathrm{CFP}}$ represents the fused cross-view features.

The process begins by aligning features $f_{\text{RGB}}$, $f_{\text{SRM}}$, $f_{\text{Bayer}}$, and $f_{\text{Noiseprint}}$ with $F_{\text{sam}}$ via bicubic interpolation. These features are then concatenated with $F_{\text{sam}}$ after a $1 \times 1$ convolution. The concatenated features pass through two stacked convolution blocks ($3 \times 3$ convolution, instance normalization, and GELU activation). Meanwhile, the semantic selection attention branch (global average pooling, fully connected layer, GELU, and sigmoid) weights the shortcut branch to produce the integrated feature $F_{\text{CFP}}$. The CFP module effectively fuses multi-view features to improve segmentation performance.

\begin{equation}
F_{\text{CFP}} = \Phi_{\text{CFP}}\left(F_{\text{sam}}, f_{\text{RGB}}, f_{\text{SRM}}, f_{\text{Bayer}},f_{\text{Noiseprint}}\right)
\end{equation}
\subsection{Prompt Mixing Module}
Since $F_{\mathrm{CFP}}$ aggregates semantically related and unrelated features, it aids the guidance of SAM for image manipulation detection. Therefore, we constructed the PMM module, which is primarily based on MLP, to integrate multiple types of prompt information. First, the two types of prompt inputs, $F_{\mathrm{CFP}}$ and $F_{\mathrm{opt}}$, are concatenated, followed by an MLP layer to perform dimension transformation to align with the original SAM prompt encoder's prompt embedding. Then, $F_{\mathrm{mix}}$ and $F_{\mathrm{CFP}}$ are input into SAM's mask decoder to achieve the image manipulation localization process. The process is as follows:
\begin{equation}
F_{\mathrm{mix}} = \Phi_{\mathrm{MLP}}\left([F_{\mathrm{CFP}} ; F_{\mathrm{opt}}]\right),
\end{equation}
\begin{equation}
P_{\mathrm{sam}} = \Phi_{\mathrm{sam-dec}}(F_{\mathrm{CFP}}, F_{\mathrm{mix}}),
\end{equation}
where $\Phi_{\mathrm{MLP}}$ represents the MLP layer operation, $F_{\mathrm{mix}}$ represents the integrated prompt embedding, and $\Phi_{\mathrm{sam-dec}}$ represents SAM's mask decoder, with $P_{\mathrm{sam}}$ representing the mask probability prediction output of the mask decoder.

\subsection{Training and Inference Pipeline}
The training loss function of IMDPrompter includes the segmentation loss from four prompt views $\mathcal{L}_{\text{Seg-p}}$, the CPC loss $\mathcal{L}_{\mathrm{CPC}}$, the SAM decoder segmentation loss $\mathcal{L}_{\text{Seg-sam}}$, and the image-level prediction loss $\mathcal{L}_{\text{Img-level}}$. The formula is expressed as follows:
\begin{equation}
\mathcal{L} = \mathcal{L}_{\text{Seg-sam}} + \lambda_1 \mathcal{L}_{\text{Seg-p}} + \lambda_2 \mathcal{L}_{\mathrm{CPC}} + \lambda_3 \mathcal{L}_{\text{Img-level}},
\end{equation}
where
\begin{equation}
L_{\text{Seg-sam}} = L_{\text{Seg}}(P_{\text{sam}}, G),
\end{equation}
\begin{equation}
L_{\text{Seg-p}} = L_{\text{Seg}}(P_{\text{RGB}}, G) + L_{\text{Seg}}(P_{\text{Bayar}}, G) + L_{\text{Seg}}(P_{\text{SRM}}, G)+ L_{\text{Seg}}(P_{\text{Noiseprint}}, G).
\end{equation}

In this paper, we use Focal Loss as our $L_{\mathrm{Seg}}$.

For image-level detection, following the work of \cite{zhai2023towards}, we adopt an adaptive pooling based on minimizing intra-class prediction variance. The overall computational process is as follows, first using Otsu's method \cite{otsu1975threshold} to find a threshold $\omega_0$ that minimizes the intra-class prediction variance:
\begin{equation}
\begin{array}{r}
\omega_0=\underset{\omega \in\left\{\hat{p}_{i, j}\right\}}{\arg \min }\left|\left\{\hat{p}_{i, j} \mid \hat{p}_{i, j}<\omega\right\}\right| \operatorname{var}\left(\left\{\hat{p}_{i, j} \mid \hat{p}_{i, j}<\omega\right\}\right)+ \\
\left|\left\{\hat{p}_{i, j} \mid \hat{p}_{i, j} \geq \omega\right\}\right| \operatorname{var}\left(\left\{\hat{p}_{i, j} \mid \hat{p}_{i, j} \geq \omega\right\}\right),
\end{array}
\end{equation}
where $\operatorname{var}(\cdot)$ denotes variance, $\hat{p}_{i, j}$ is the pixel-level response at position $(i, j)$.
Then the image-level prediction is aggregated from pixel-level responses above the threshold and the image-level loss is:
\begin{equation}
\hat{y}_{\mathrm{A}} = \frac{1}{\left| \mathbb{P}_{\mathrm{h}} \right|} \sum_{\hat{p} \in \mathbb{P}_{\mathrm{h}}} \hat{p}; \mathbb{P}_{\mathrm{h}} = \left\{\hat{p}_{i, j} \mid \hat{p}_{i, j} \geq \omega_0\right\},
\end{equation}
\begin{equation}
\mathcal{L}_{\text{Img-level}} = \mathcal{L}_{\mathrm{BCE}}(y, \hat{y}_{\mathrm{A}}),
\end{equation}
where $\hat{y}_{\mathrm{A}}$ is the image-level prediction, $\mathcal{L}_{\mathrm{BCE}}$ denotes the binary cross-entropy loss.

During the inference process of IMDPrompter, since there are no true labels, our OPS module defaults to selecting $P_{\text{Ens}}$ for generating  masks and bounding box prompts, with other components functioning as during the training process.

\section{Experiment}

\noindent\textbf{Dataset.} Our method is trained only on the CASIAv2 dataset \cite{dong2013casia}. For in-distribution (IND) evaluation, we use the CASIAv1 dataset \cite{dong2013casia}. For out-of-distribution (OOD) evaluation, we use three datasets: Columbia \cite{hsu2006detecting}, Coverage \cite{wen2016coverage}and IMD2020 \cite{novozamsky2020imd2020}. 

\noindent\textbf{Evaluation Metrics.} For image-level manipulation detection, we report specificity, sensitivity, and their F1-score (I-F1). The area under the receiver operating characteristic curve (AUC) is also reported as a threshold-independent metric for image-level detection. For pixel-level manipulation localization, we follow previous methods \cite{chen2021image,zhou2018learning,zhou2018generate,salloum2018image} to compute pixel accuracy, recall, and their F1-score (P-F1) on manipulated images. The overall performance of image and pixel-level manipulation detection/localization is measured by the harmonic mean of pixel-level and image-level F1-scores \cite{chen2021image}, denoted as composite F1 (C-F1), and is sensitive to lower values of P-F1 and I-F1. To ensure fair comparison, a default threshold of 0.5 is used for F1 computation unless otherwise specified.

\noindent\textbf{Implementation Details.} In our experiments, unless otherwise specified, we consistently use the VIT-L backbone of SAM and employ FCN (lightweight architecture based on MobileNet \cite{howard2017mobilenets}) as the segmentor for the three prompt views. We maintain image size at 1024×1024, consistent with the original input of the SAM model. To augment training samples, we use data augmentation techniques such as horizontal flipping and random cropping. The image encoder remains frozen during the training phase. All experiments are run on NVIDIA A6000 GPUs. For the optimization process, we train our model using the AdamW optimizer with an initial learning rate of 1e-4. We use a batch size of 4 and train for 100 epochs. We implement a linear warm-up strategy with a cosine annealing scheduler \cite{loshchilov2016sgdr} to decay the learning rate. 

\subsection{Comparison with State-of-the-Art Method}

\begin{table*}[ht]
\centering
\caption{Pixel-level manipulation detection performance. Performance metric is F1-score (\%). The best results for each test set are highlighted in bold and second-best values are underlined. }
\renewcommand\arraystretch{1.2}
\resizebox{\linewidth}{!}{

\begin{tabular}{l|cccccc|cccccc}
\hline
                                                      & \multicolumn{6}{c|}{Best threshold}                                                                                                                                                     & \multicolumn{6}{c}{Fixed threshold (0.5)}                                                                                                                                               \\ \cline{2-13} 
\multirow{-2}{*}{Method}                              & CASIA                        & COVER                        & Columbia                     & IMD                          & NIST                         & MEAN                         & CASIA                        & COVER                        & Columbia                     & IMD                         & NIST                         & MEAN                          \\ \hline
MFCN \cite{salloum2018image}         & 54.1                         & -                            & 61.2                         & -                            & -                            & -                            & -                            & -                            & -                            & -                           & -                            & -                             \\
RGB-N \cite{zhou2018learning}        & 40.8                         & 37.9                         & -                            & -                            & -                            & -                            & -                            & -                            & -                            & -                           & -                            & -                             \\
H-LSTM \cite{bappy2019hybrid}        & 20.9                         & 21.3                         & 14.2                         & 31.0                         & 46.6                         & 26.80                        & 15.4                         & 16.3                         & 13.0                         & 19.5                        & 35.4                         & 19.92                         \\
ManTra-Net \cite{wu2019mantra}       & 69.2                         & 77.2                         & 70.9                         & 70.5                         & 45.5                         & 66.66                        & 15.5                         & 28.6                         & 36.4                         & 18.7                        & 0.0                          & 19.84                         \\
HP-FCN \cite{li2019localization}     & 21.4                         & 19.9                         & 47.1                         & 16.9                         & 36.0                         & 28.26                        & 15.4                         & 0.3                          & 6.7                          & 11.2                        & 12.1                         & 9.14                          \\
CR-CNN \cite{yang2020constrained}    & 66.2                         & 47.0                         & 70.4                         & 60.0                         & 42.8                         & 57.28                        & 40.5                         & 29.1                         & 43.6                         & \underline{26.2}                        & 23.8                         & 32.64                         \\
GSR-Net \cite{zhou2018generate}      & 57.4                         & 48.9                         & 62.2                         & 68.7                         & 45.6                         & 56.56                        & 38.7                         & 28.5                         & 61.3                         & 24.3                        & 28.3                         & 36.22                         \\
SPAN \cite{hu2020span}               & 68.8                         & 71.8                         & 77.4                         & 69.6                         & 68.3                         & 71.18                        & 18.4                         & 17.2                         & 48.7                         & 17.0                        & 22.1                         & 24.68                         \\
CAT-Net \cite{kwon2021cat}           & 57.3                         & 48.5                         & 77.6                         & 51.7                         & 59.9                         & 59.00                        & 13.6                         & 12.9                         & 55.5                         & 5.4                         & 17.9                         & 21.06                         \\
MVSS-Net \cite{chen2021image}        & \underline{75.3}            & \underline{82.4}            & 70.3                         & \underline{75.7}            & \underline{73.7}            & \underline{75.48}           & 45.2                         & 45.3                         & 63.8                         & 26.0                        & 29.2                         & \underline{41.90}                          \\
Trufor \cite{guillaro2023trufor}     & 83.5                         & 74.1                         & \underline{82.1}                          & -                            & 68.8                         & -                            & \underline{73.7}                         & \underline{60.0}                        & \underline{85.9}                        & -                           & \underline{39.9}                          & -                             \\
\rowcolor[HTML]{E8E6E6} 
FCN \cite{long2015fully}             & 74.2                         & 57.3                         & 58.6                         & 64.5                         & 50.7                         & 61.06                        & 44.1                         & 19.9                         & 22.3                         & 21.0                        & 16.7                         & 24.80                         \\
\rowcolor[HTML]{E8E6E6} 
\cellcolor[HTML]{E8E6E6}                              & \textbf{85.1}                & \textbf{83.5}                & \textbf{87.3}                & \textbf{76.3}                & \textbf{74.6}                & \textbf{81.36}               & \textbf{76.3}                & \textbf{63.6}                & \textbf{87.3}                & \textbf{30.6}               & \textbf{41.1}                & \textbf{59.78}                \\
\rowcolor[HTML]{E8E6E6} 
\multirow{-2}{*}{\cellcolor[HTML]{E8E6E6}IMDPrompter} & {\color[HTML]{FE0000} +10.9} & {\color[HTML]{FE0000} +26.2} & {\color[HTML]{FE0000} +28.7} & {\color[HTML]{FE0000} +11.8} & {\color[HTML]{FE0000} +23.9} & {\color[HTML]{FE0000} +20.3} & {\color[HTML]{FE0000} +32.2} & {\color[HTML]{FE0000} +43.7} & {\color[HTML]{FE0000} +65.0} & {\color[HTML]{FE0000} +9.6} & {\color[HTML]{FE0000} +24.4} & {\color[HTML]{FE0000} +34.98} \\ \hline
\end{tabular}

}
\label{4-1-1}
\end{table*}

\begin{table*}[ht]
\centering
\caption{Image-level manipulation detection performance. Decision threshold: 0.5. NIST16 is excluded as it lacks genuine images. The best results for each test set are highlighted in bold and second-best values are underlined.}
\renewcommand\arraystretch{1.2}
\resizebox{\linewidth}{!}{

\begin{tabular}{l|cccc|cccc|cccc|cccc|cc}
\hline
                                                      & \multicolumn{4}{c|}{CASIA}                                                                                                & \multicolumn{4}{c|}{COVER}                                                                                                & \multicolumn{4}{c|}{Columbia}                                                                                             & \multicolumn{4}{c|}{IMD}                                                                                                   & \multicolumn{2}{c}{MEAN}                                       \\ \cline{2-19} 
\multirow{-2}{*}{Method}                              & AUC                           & Sen.                         & Spe.                         & F1                          & AUC                           & Sen.                        & Spe.                         & F1                           & AUC                           & Sen.                        & Spe.                         & F1                           & AUC                           & Sen.                         & Spe.                         & F1                           & AUC                            & F1                            \\ \hline
H-LSTM \cite{bappy2019hybrid}        & 0.498                         & 99.7                         & 0.0                          & 0.0                         & 0.500                         & 100.0                       & 0.0                          & 0.0                          & 0.506                         & 100.0                       & 1.1                          & 2.2                          & 0.501                         & 100.0                        & 0.0                          & 0.0                          & 0.5013                         & 0.55                          \\
ManTra-Net \cite{wu2019mantra}       & 0.500                         & 100.0                        & 0.0                          & 0.0                         & 0.500                         & 100.0                       & 0.0                          & 0.0                          & 0.701                         & 100.0                       & 0.0                          & 0.0                          & 0.500                         & 100.0                        & 0.0                          & 0.0                          & 0.5503                         & 0.00                          \\
CR-CNN \cite{yang2020constrained}    & 0.719                         & 93.0                         & 13.9                         & 24.2                        & 0.566                         & 96.7                        & 7.0                          & 13.1                         & 0.783                         & 96.1                        & 24.6                         & 39.2                         & 0.615                         & 92.9                         & 12.3                         & 21.7                         & 0.6708                         & 24.55                         \\
GSR-Net \cite{zhou2018generate}      & 0.500                         & 99.4                         & 0.0                          & 0.0                         & 0.515                         & 100.0                       & 0.0                          & 0.0                          & 0.502                         & 100.0                       & 1.1                          & 2.2                          & 0.500                         & 100.0                        & 0.0                          & 0.0                          & 0.5043                         & 0.55                          \\
SPAN \cite{hu2020span}               & 0.500                         & 100.0                        & 0.0                          & 0.0                         & 0.500                         & 100.0                       & 0.0                          & 0.0                          & 0.500                         & 100.0                       & 0.0                          & 0.0                          & 0.500                         & 100.0                        & 0.0                          & 0.0                          & 0.5000                         & 0.00                          \\
CAT-Net \cite{kwon2021cat}           & 0.647                         & 23.9                         & 92.1                         & 38.0                        & 0.557                         & 28.0                        & 80.0                         & \underline{41.5}                            & 0.971                         & 87.2                        & 96.2                         & \underline{91.5}                            & 0.586                         & 27.5                         & 81.6                         & \underline{41.1}                            & 0.6903                         & 53.03                         \\
MVSS-Net \cite{chen2021image}        & \underline{0.937}                            & 61.5                         & 98.8                         & \underline{75.8}                           & 0.731                         & 94.0                        & 14.0                         & 24.4                         & 0.980                         & 66.9                        & 100.0                        & 80.2                         & \underline{0.656}                            & 91.5                         & 22.0                         & 35.5                         & \underline{0.8260}                            & \underline{53.98}                            \\
Trufor \cite{guillaro2023trufor}     & 0.916                         & -                            & -                            & -                           &   \underline{0.770}                         & -                           & -                            & -                            & \textbf{0.996 }                        & -                           & -                            & -                            & -                             & -                            & -                            & -                            & -                              & -                             \\
\rowcolor[HTML]{E8E6E6} 
FCN \cite{long2015fully}             & 0.770                            & 72.8                         & 64.3                         & 68.3                        & 0.541                         & 90.0                        & 10.0                         & 18.0                         & 0.762                         & 95.0                        & 32.2                         & 48.1                         & 0.502                         & 84.6                         & 15.5                         & 26.2                         & 0.6438                         & 40.15                         \\
\rowcolor[HTML]{E8E6E6} 
\cellcolor[HTML]{E8E6E6}                              &\textbf{ 0.978  }                       & 91.6                         & 66.9                         & \textbf{77.3 }                       & \textbf{0.796 }                        & 96.8                        & 55.2                         &\textbf{ 70.3}                         & \underline{0.983}                            & 97.3                        & 90.2                         & \textbf{93.6 }                        & \textbf{0.671}                         & 71.6                         & 57.4                         &\textbf{ 63.7}                         & \textbf{0.8570 }                        & \textbf{76.23 }                           \\
\rowcolor[HTML]{E8E6E6} 
\multirow{-2}{*}{\cellcolor[HTML]{E8E6E6}IMDPrompter} & {\color[HTML]{FE0000} +0.208} & {\color[HTML]{FE0000} +18.8} & {\color[HTML]{FE0000} +2.6} & {\color[HTML]{FE0000} +9.0} & {\color[HTML]{FE0000} +0.255} & {\color[HTML]{FE0000} +6.8} & {\color[HTML]{FE0000} +45.2} & {\color[HTML]{FE0000} +52.3} & {\color[HTML]{FE0000} +0.221} & {\color[HTML]{FE0000} +2.3} & {\color[HTML]{FE0000} +58.0} & {\color[HTML]{FE0000} +45.5} & {\color[HTML]{FE0000} +0.169} & {\color[HTML]{009901} -13.0} & {\color[HTML]{FE0000} +41.9} & {\color[HTML]{FE0000} +37.5} & {\color[HTML]{FE0000} +0.2132} & {\color[HTML]{FE0000} +36.08} \\ \hline
\end{tabular}

}
\label{4-1-2}
\end{table*}

\begin{figure*}[t]
	\centering
	\includegraphics[width=0.85\linewidth]{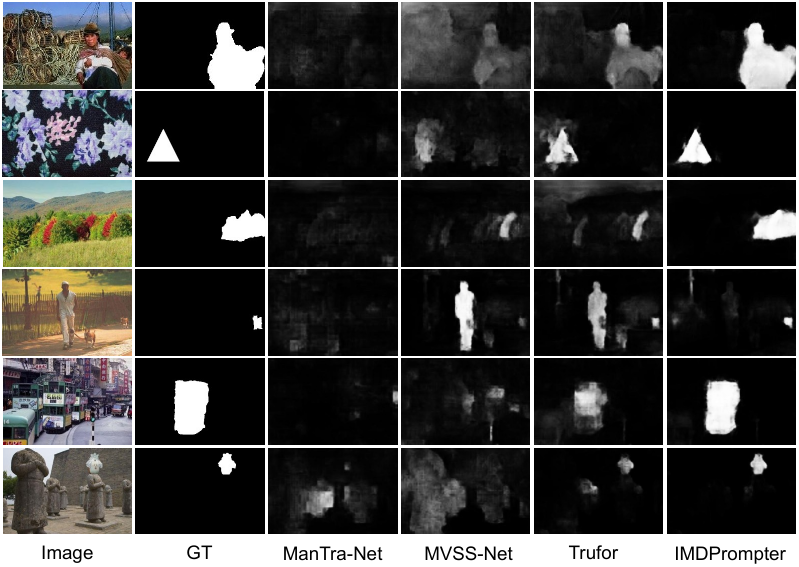}
	\caption{ Visualization Comparison of Image Manipulation Detection and Localization Results}
	\label{vis}
\end{figure*}

\begin{table}[htbp]
\begin{minipage}[t]{0.48\textwidth} % 左侧半页
\centering
\caption{Evaluation of each component of our method in IND dataset.} % 标题
\renewcommand\arraystretch{1.2}
\resizebox{\columnwidth}{!}{
\begin{tabular}{c|cccc|ccc}
\hline
\multirow{2}{*}{} & \multirow{2}{*}{CFP} & \multirow{2}{*}{CPC} & \multirow{2}{*}{OPS} & \multirow{2}{*}{PMM} & \multicolumn{3}{c}{CASIA} \\ \cline{6-8} 
&&& && I-F1 & P-F1 & C-F1  \\ \hline
{1} &&& && 70.6 & 70.3 & 70.4    \\
{2} & $\checkmark$ && && 73.2 & 73.1 & 73.1    \\
{3}& $\checkmark$ & $\checkmark$ & && 74.3 & 74.3 & 74.3    \\
{4} & $\checkmark$ & $\checkmark$ & $\checkmark$&& 75.1 & 76.1 & 75.6    \\
{5} & $\checkmark$ & $\checkmark$ & $\checkmark$& $\checkmark$ & 77.3 & 76.3 & 76.8    \\ \hline
\end{tabular}
}
\label{AS1}
\end{minipage}%
\hspace{0.3cm}  % 添加空格以隔开两个 minipage
\begin{minipage}[t]{0.48\textwidth} % 右侧半页
\centering
\caption{Evaluation of each component of our method in OOD dataset.} % 标题
\renewcommand\arraystretch{1.2}
\resizebox{\columnwidth}{!}{
\begin{tabular}{c|cccc|ccc}
\hline
\multirow{2}{*}{} & \multirow{2}{*}{CFP} & \multirow{2}{*}{CPC} & \multirow{2}{*}{OPS} & \multirow{2}{*}{PMM} & \multicolumn{3}{c}{COVER} \\ \cline{6-8} 
&&& && I-F1 & P-F1 & C-F1  \\ \hline
{6} &&& && 20.4 & 39.8 & 27.0    \\
{7} & $\checkmark$ && && 49.6 & 54.3 & 51.8    \\
{8} & $\checkmark$ & $\checkmark$ & && 58.6 & 57.0 & 57.8    \\
{9} & $\checkmark$ & $\checkmark$ & $\checkmark$&& 65.2 & 61.4 & 63.2    \\
{10}& $\checkmark$ & $\checkmark$ & $\checkmark$& $\checkmark$ & 70.3 & 63.6 & 66.8    \\ \hline
\end{tabular}
}
\label{AS2}
\end{minipage}
\end{table}

\begin{table}[h]
\centering
\caption{Ablation Study of View Combinations}
\resizebox{\columnwidth}{!}{
\begin{tabular}{cccc|cccc|cccc|cccc|cccc}
\hline
\multirow{2}{*}{RGB} & \multirow{2}{*}{Noiseprint} & \multirow{2}{*}{Bayar} & \multirow{2}{*}{SRM} & \multicolumn{4}{c|}{CASIA}& \multicolumn{4}{c|}{COVER}& \multicolumn{4}{c|}{Columbia} & \multicolumn{4}{c}{IMD}       \\ \cline{5-20} 
 &       &  && I-AUC & I-F1& P-F1& C-F1& I-AUC & I-F1& P-F1& C-F1& I-AUC & I-F1& P-F1& C-F1& I-AUC & I-F1& P-F1& C-F1  \\ \hline
\checkmark&       &  && 0.856 & 73.60 & 69.60 & 71.50 & 0.556 & 19.60 & 38.60 & 26.00 & 0.786 & 53.60 & 24.90 & 34.00 & 0.531 & 28.10 & 21.20 & 24.20 \\
 & \checkmark     &  && 0.903 & 75.90 & 73.60 & 74.70 & 0.761 & 62.90 & 58.70 & 60.70 & 0.946 & 88.60 & 83.50 & 86.00 & 0.636 & 59.90 & 27.90 & 38.10 \\
 &       & \checkmark&& 0.863 & 74.30 & 70.60 & 72.40 & 0.706 & 61.30 & 56.30 & 58.70 & 0.889 & 83.40 & 81.60 & 82.50 & 0.601 & 54.60 & 25.70 & 34.90 \\
 &       &  & \checkmark& 0.896 & 75.10 & 71.60 & 73.30 & 0.713 & 59.60 & 55.60 & 57.50 & 0.906 & 84.60 & 79.30 & 81.90 & 0.593 & 56.30 & 24.30 & 33.90 \\
\checkmark& \checkmark     &  && 0.941 & 76.90 & 75.20 & 76.00 & 0.781 & 66.80 & 61.90 & 64.30 & 0.969 & 91.70 & 85.40 & 88.40 & 0.655 & 61.50 & 28.80 & 39.20 \\
\checkmark& \checkmark     & \checkmark&& 0.962 & 77.10 & 75.90 & 76.50 & 0.788 & 68.70 & 62.40 & 65.40 & 0.977 & 92.40 & 86.50 & 89.40 & 0.660 & 62.60 & 29.50 & 40.10 \\
\checkmark& \checkmark     & \checkmark& \checkmark& 0.978 & 77.30 & 76.30 & 76.80 & 0.796 & 70.30 & 63.60 & 66.80 & 0.983 & 93.60 & 87.30 & 90.30 & 0.671 & 63.70 & 30.60 & 41.30 \\ \hline
\end{tabular}
}
\label{tab:view-com}
\end{table}

% Please add the following required packages to your document preamble:
% \usepackage{multirow}
\begin{table}[h]
\caption{Ablation Study of SAM}
\resizebox{\textwidth}{!}{
\begin{tabular}{c|cccc|cccc|cccc|cccc}
\hline
\multirow{2}{*}{Method} & \multicolumn{4}{c|}{CASIA} & \multicolumn{4}{c|}{COVER} & \multicolumn{4}{c|}{Columbia} & \multicolumn{4}{c}{IMD}    \\ \cline{2-17} 
  & I-AUC & I-F1 & P-F1 & C-F1 & I-AUC & I-F1 & P-F1 & C-F1 & I-AUC& I-F1& P-F1& C-F1 & I-AUC & I-F1 & P-F1 & C-F1 \\ \hline
FCN & 0.770& 68.3 & 44.1 & 53.6 & 0.541 & 18.0 & 19.9 & 18.9 & 0.762& 48.1& 22.3& 30.5 & 0.502 & 26.2 & 21.0 & 23.3 \\
FCN+& 0.873 & 73.4 & 61.3 & 66.8 & 0.681 & 49.2 & 43.9 & 46.4 & 0.871& 72.6& 52.4& 60.9 & 0.589 & 43.1 & 25.1 & 31.7 \\
IMDPrompter & 0.978 & 77.3 & 76.3 & 76.8 & 0.796 & 70.3 & 63.6 & 66.8 & 0.983& 93.6& 87.3& 90.3 & 0.671 & 63.7 & 30.6 & 41.3 \\ \hline
\end{tabular}
}
\label{tab:sam}
\end{table}

\begin{figure}[h]
	\centering
 	\includegraphics[height=9.5cm,width=12cm]{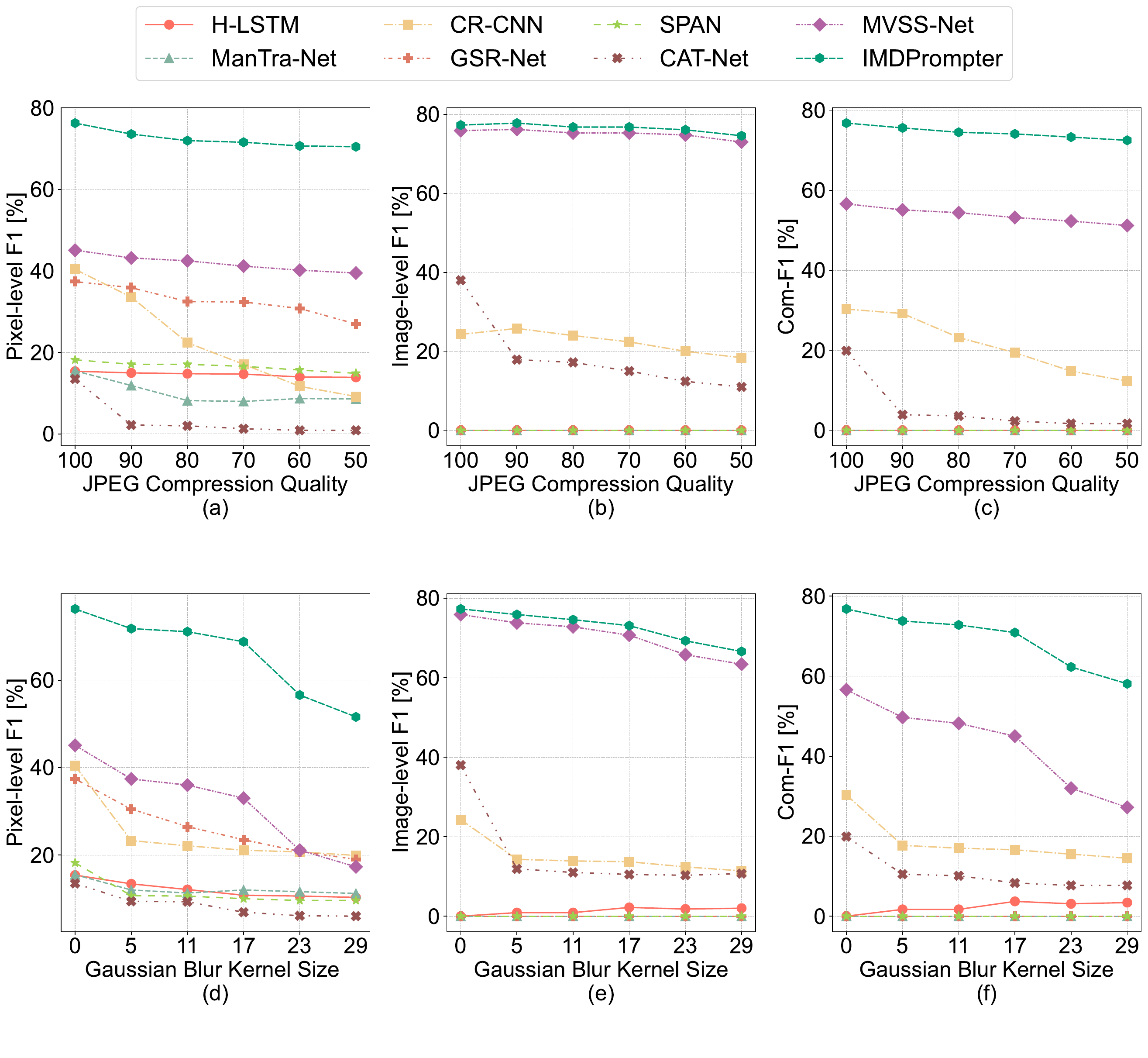}

	\caption{Robustness Analysis Against JPEG Compression and Gaussian Blur Interference}
 
	\label{sensitivitybar}
\end{figure}

\noindent\textbf{Pixel-level Manipulation Detection}
Table \ref{4-1-1} illustrates the pixel-level detection performance of different models. We evaluate the F1-score under two settings: best threshold and fixed threshold (0.5). From Table \ref{4-1-1}, it is evident that IMDPrompter achieves the best performance across nearly all datasets. Under the optimal threshold setting, we achieved an average F1 Score of 81.36\%. Under the fixed threshold setting, IMDPrompter achieved an average F1 Score of 59.78\%, indicating that our proposed IMDPrompter demonstrates better robustness in threshold settings.

\noindent\textbf{Image-level Manipulation Detection.}
Table \ref{4-1-2} presents the image-level performance of different models. For image-level performance, we use a default decision threshold of 0.5. Once again, IMDPrompter emerges as the top performer, leveraging our prompt learning paradigm to achieve higher specificity in most test settings, thereby reducing false positives. Additionally, IMDPrompter achieves the best average AUC and average F1-scores. The average F1-score of IMDPrompter significantly surpasses that of the second-best method, MVSS-Net, with an improvement of 22.25\%.

\noindent\textbf{Qualitative Results.}
As shown in Figure \ref{vis}, we visualized the image manipulation localization results of four methods: ManTra-Net, MVSS-Net, Trufor, and IMDPrompter. It is evident that IMDPrompter generates more accurate confidence maps for image manipulation localization, achieving better detection and localization.

\subsection{Ablation Studies}

To reveal the impact of different components, we evaluated the proposed model under various settings, analyzing the distinct effects of each component on in-domain and out-of-domain datasets. As shown in Tables \ref{AS1} and \ref{AS2}, we conducted ablation experiments on the in-domain dataset CASIA and the out-of-domain dataset COVER, respectively.

\noindent\textbf{Baseline.}  From \textit{Experiment {1}}, it is known that our baseline method achieved performance scores of 70.6\%, 70.3\%, and 70.4\% on the I-F1, P-F1, and C-F1 metrics respectively in the CASIA dataset. In the COVER dataset, it scored 20.4\%, 39.8\%, and 27.0\% on the same metrics. These experiments show that using only RGB visual view information has limited generalization ability in out-of-domain datasets.

\noindent\textbf{Impact of CFP.} From \textit{Experiment {2}}, By implementing cross-view feature perception fusion through the CFP module, the I-F1, P-F1, and C-F1 metrics on the CASIA dataset improved by 2.6\%, 2.8\%, and 2.7\%, respectively. Experiment VII showed that on the COVER dataset, the I-F1, P-F1, and C-F1 metrics increased by 29.2\%, 14.5\%, and 24.8\%, respectively. This indicates that cross-view feature perception fusion can enhance performance on both in-domain and out-of-domain datasets, especially on out-of-domain datasets.

\noindent\textbf{Impact of CPC.} From \textit{Experiment {3}}, the inclusion of CPC, which introduced cross-view consistency enhancement, resulted in performance gains of 3.7\%, 4.0\%, and 3.9\% on the I-F1, P-F1, C-F1 metrics respectively in the CASIA dataset. Experiment VIII showed performance gains of 38.2\%, 17.2\%, and 30.8\% respectively on the COVER dataset, confirming that the prompt information between different views is complementary and integrative enhancement can bring objective performance improvements.

\noindent\textbf{Impact of OPS.} The inclusion of OPS for heuristic selection of optimal prompts resulted in improvements of 4.5\%, 5.8\%, and 5.2\% in the I-F1, P-F1, and C-F1 metrics on the CASIA dataset, respectively. Experiment IX showed gains of 44.8\%, 21.6\%, and 36.2\% respectively on the COVER dataset, indicating that simply adding together prompts from different views is not enough to generate the best prompts, and a optimal prompts selection strategy can optimize the prompt selection process.
From \textit{Experiment {4}}, the inclusion of OPS adaptive selection of the optimal prompts resulted in performance gains of 4.5\%, 5.8\%, and 5.2\% on the I-F1, P-F1, C-F1 metrics respectively in the CASIA dataset. 

\noindent\textbf{Impact of PMM.} From \textit{Experiment {5}}, with the inclusion of PMM which fully integrates various types of prompt information, there was an increase in performance of 6.7\%, 6.0\%, and 6.4\% on the I-F1, P-F1, C-F1 metrics respectively in the CASIA dataset. \textit{Experiment {10}} showed an increase of 49.9\%, 23.8\%, and 39.8\% respectively on the COVER dataset, further demonstrating the effectiveness of PMM.

\noindent\textbf{Impact of View Combinations.} 
As shown in Table \ref{tab:view-com}, the performance of using the RGB view alone does not differ significantly from that of using semantic-agnostic views such as SRM, Bayer, and Noiseprint individually on the in-domain test set CASIA.
In contrast, for the out-of-domain test sets COVER, Columbia, and IMD, there is a significant performance gap between the RGB view and the semantic-agnostic views. This indicates that semantic-agnostic views play a crucial role in generalization on out-of-domain datasets. Additionally, we conducted ablation studies on various view combinations, and using both the RGB view and the Noiseprint view simultaneously resulted in substantial performance improvements. This demonstrates that the Noiseprint view serves as a critical semantic-agnostic feature. Next, we sequentially introduced the Bayer and SRM views, achieving further steady improvements across all metrics. When we combined the RGB, Noiseprint, SRM, and Bayer views, our method achieved optimal performance, validating the effectiveness of this combination.

\noindent\textbf{Impact of SAM.} As shown in Table \ref{tab:sam}, we constructed an FCN+ model that integrates four views without using SAM, further analyzing the impact of SAM on IMD. Due to the lack of potential priors from a large-scale pre-training dataset, the performance ceiling of FCN+ is relatively low. In contrast, our IMDPrompter fully leverages these potential priors, thereby enhancing the performance ceiling

\noindent\textbf{Impact of Quality Degradation.} As shown in Figure \ref{sensitivitybar}, following \cite{dong2022mvss}, we evaluated the robustness of the models under two common image processing operations encountered during the dissemination of images on the internet, namely JPEG compression and Gaussian blur. Comparing these two operations, Gaussian blur has a more significant impact on detection performance, especially when using larger 17x17 scale convolution kernels. Compared to previous methods, our IMDPrompter exhibits better robustness.

\section{Conslusion}
In this paper, we propose a cross-view perceptual prompt learning paradigm, IMDPrompter, which applies SAM to the image manipulation detection task for the first time. We sequentially introduce components such as optimal prompt selection, cross-view prompt consistency, cross-view feature perception, and prompt mixing modules to achieve efficient and automated image manipulation detection and localization. Our IMDPrompter demonstrates significant image-level and pixel-level manipulation detection performance in both in-domain (IND) and out-of-domain (OOD) scenarios, as well as across various robustness evaluation settings.

\section{ACKNOWLEDGEMENTS}
We thank all the anonymous reviewers for their helpful comments.
This work was supported by the National Key R\&D Program of China (2022YFB4701400/4701402), SSTIC Grant(KJZD20230923115106012,KJZD20230923114916032,GJHZ20240218113604008), Beijing Key Lab of Networked Multimedia and National Natural Science Foundation of China under Grant 62202302.
\newpage
% \clearpage

\bibliographystyle{iclr2025_conference}
\bibliography{iclr2025_conference}

\clearpage

\section{Appendix}
\subsection{Limitations and Broader Impacts}
\textbf{Limitations. }We have some limitations in our method. First, it cannot detect completely generated images. Second, training IMDPrompter requires complete pixel-level supervision. In future research, we plan to explore more efficient label utilization through weakly supervised and semi-supervised learning approaches and extend these methods to newer foundational models such as SEEM \cite{zou2024segment} and SAM2 \cite{ravi2024sam}. \\
\textbf{Broader Impacts. }In recent years, with the advancement of deep learning technologies \cite{
zhang2024distilling,
zhang2025rethinking,
qiu2024tfb,
qiu2025duet,
qiu2025easytime,
wu2024catch,
AutoCTS++,
li2024foundts,
hu2024multirc,
wei2024free,
wei2024task,
wei2025modeling,
zhang2024can,
dong2023dfvsr,
dong2023enhanced,
dong2023frequency}, the editing and manipulation of digital media have become increasingly accessible and widespread. Advances in image editing software, along with deep generative models such as Generative Adversarial Networks and diffusion models \cite{
gao2024diffimp,
chen2024gim}, have made image manipulations, often imperceptible to the human eye, much easier. These technologies are even being widely employed by potential malicious users. The ubiquitous use of smartphones and social networks has further accelerated the dissemination of these manipulated media. As a result, when these edited images are employed to support disinformation or to distort news content in order to mislead the public, they can create significant social problems and contribute to a crisis of trust. In this context, IMDPrompter, as an effective image manipulation detection method, can play a crucial role in mitigating the negative consequences of intentional image manipulation.

\subsection{Dataset Description}
In order to directly compare with state-of-the-art technologies, we trained on CASIAv2 \cite{dong2013casia} and conducted extensive testing on COVER \cite{wen2016coverage}, Columbia \cite{hsu2006detecting}, NIST16 \cite{hsu2006detecting}, CASIAv1 \cite{dong2013casia}, and the recent IMD \cite{novozamsky2020imd2020}.
\begin{table}[ht]
\centering
\renewcommand{\arraystretch}{1.2} % Adjusts the row height
\setlength{\tabcolsep}{5pt} % Adjusts the space between columns
\caption{Details of the training set and five test sets used in our experiments. The symbol “–” indicates unavailable information. Copy-move, splicing, and inpainting operations are denoted as cpmv, spli, and inpa, respectively. Our model was trained on the CASIAv2 dataset and evaluated across all test sets.}
\begin{tabular}{lcccccc}
\toprule
\textbf{Dataset} & \textbf{Negative} & \textbf{Positive} & \textbf{cpmv} & \textbf{spli} & \textbf{inpa} \\
\midrule
\textbf{Training} & & & & & \\
CASIAv2 \cite{dong2013casia} & 7491 & 5063 & 3235 & 1828 & 0 \\
\textbf{Testing} & & & & & \\
COVER \cite{wen2016coverage} & 100 & 100 & 100 & 0 & 0 \\
Columbia  \cite{hsu2006detecting} & 183 & 180 & 0 & 180 & 0 \\
NIST16 \cite{dong2013casia} & 0 & 564 & 68 & 288 & 208 \\
CASIAv1+ \cite{dong2013casia} & 800 & 920 & 459 & 461 & 0 \\
IMD \cite{novozamsky2020imd2020} & 414 & 2010 & -- & -- & -- \\
\bottomrule
\end{tabular}
\label{tab:dataset_details}
\end{table}
\subsection{Baseline Methods}

% Please add the following required packages to your document preamble:
% \usepackage{multirow}
% \usepackage[table,xcdraw]{xcolor}
% Beamer presentation requires \usepackage{colortbl} instead of \usepackage[table,xcdraw]{xcolor}
\begin{table}[h]
\centering
\caption{Comparison of Technical Routes of Baseline Methods}
\resizebox{\linewidth}{!}{
\begin{tabular}{c|ccc|c}
\hline
   & \multicolumn{3}{c|}{View}                                                                                                                  & Semantic Segmentation \\ \cline{2-4}
\multirow{-2}{*}{Method} & RGB& Noise& Fusion                                                                                   & Backbone              \\ \hline
MFCN\cite{salloum2018image} & +  & -& -                                                                                        & FCN                   \\
RGB-N\cite{zhou2018learning}& +  & SRM& Late Fusion(Bilinear Pooling)                                                            & Faster R-CNN          \\
H-LSTM\cite{bappy2019hybrid} & +  & -& -                                                                                        & Patch-LSTM            \\
ManTra-Net\cite{wu2019mantra} & +  & SRM+Bayar& {\color[HTML]{000000} Early Fusion(Feature Concatenation)}                               & Wider VGG             \\
HP-FCN\cite{li2019localization} & -  & High-pass filters& -                                                                                        & FCN                   \\
GSR-Net\cite{zhou2018generate}& +  & -& -                                                                                        & DeepLabv2             \\
CR-CNN\cite{yang2020constrained} & -  & Bayar& -                                                                                        & Mask R-CNN            \\
SPAN\cite{hu2020span} & +  & SRM+Bayar& Early Fusion(Feature Concatenation)                                                      & Wider VGG             \\
MM-Net\cite{yang2021multi} & +  & Bayar& Middle Fusion(Attention Guidance)                                                        & Mask R-CNN            \\
JPEG-ComNet\cite{rao2021self}& +  & SRM& Early Fusion(Feature Concatenation)                                                      & Siamese FCN           \\
CAT-Net\cite{kwon2021cat}& +  & DCT& {\color[HTML]{000000} Middle Fusion(Feature Concatenation)}                              & HRNet                 \\
MVSS-Net\cite{chen2021image} & +  & Bayar& Late Fusion(Dual Attention)                                                              & FCN                   \\
TruFor \cite{guillaro2023trufor}& +  & Noiseprint & Late Fusion(Feature Concatenation)                                                       & Segformer             \\ \hline
IMDPrompter& {\color[HTML]{09008B} +} & SRM+Bayar+Noiseprint & \begin{tabular}[c]{@{}c@{}}Late Fusion(Optimal Prompt Selection\\ +Cross-view Prompt Consistency)\end{tabular} & FCN+SAM               \\ \hline
\end{tabular}
}
\label{tab:baseline}
\end{table}

As shown in Table \ref{tab:baseline}, we have organized the technical routes of the baseline methods. To ensure fair and reproducible comparisons, we selected state-of-the-art models that meet any of the following criteria: 1) pretrained models released by the paper authors, 2) publicly available source code, and 3) adherence to a common evaluation protocol, where CASIAv2 is used for training and other public datasets for testing. Based on these criteria, we compiled a list of nine published baselines, as follows:
\begin{itemize}
    \item H-LSTM \cite{bappy2019hybrid}: Pretrained on a custom dataset of 65k processed images and fine-tuned on NIST16 and IEEE Forensics Challenge data.
    \item ManTra-Net \cite{wu2019mantra}: Trained on millions of processed images in a private collection.
    \item HP-FCN \cite{li2019localization}: Trained on a private set of repaired images.
    \item CR-CNN \cite{yang2020constrained}: Trained on CASIAv2.
    \item SPAN \cite{hu2020span}: Trained using the same data as ManTra-Net and fine-tuned on CASIAv2.
    \item CAT-Net \cite{kwon2021cat}: Trained on a joint dataset including CASIAv2, IMD, Fantastic Reality \cite{kniaz2019point}, and self-spliced COCO.
\end{itemize}
For models with publicly available code, we trained them using the code provided by the authors, such as GSR-Net \cite{zhou2018generate}.
When citing their results, we adopt the original data where appropriate and only use our retrained models when necessary. For models that follow the same evaluation protocol, such as MFCN \cite{salloum2018image} and RGB-N \cite{zhou2018learning}, we directly cite results from the same team \cite{yang2020constrained}.
For fair comparison, we retrained FCN \cite{long2015fully}, MVSS-Net\cite{dong2022mvss}, and Trufor \cite{guillaro2023trufor} from scratch on CASIAv2. As previous studies rarely report their image-level performance, these models typically lack an image classification head in their implementations. To obtain a baseline for image-level predictions without modifying their models or code, we adopted the GMP strategy used in MVSS-Net.

\begin{table}[ht]
\centering
\caption{Parameter Settings Details}
\renewcommand{\arraystretch}{1.3} 
\label{supp-detail}
\resizebox{0.5\linewidth}{!}{
\begin{tabular}{ll}
\toprule
\textbf{Parameter} & \textbf{Value} \\
\midrule
\multicolumn{2}{l}{\textbf{Warm-up Learning Rate Scheduler}} \\
Type & LinearLR \\
Start Factor & 1.00E-04 \\
By Epoch & TRUE \\
Begin & 0 \\
End & 1 \\
Convert to Iter Based & TRUE \\
\midrule
\multicolumn{2}{l}{\textbf{Main Learning Rate Scheduler}} \\
Type & CosineAnnealingLR \\
T\_max & max\_epochs \\
By Epoch & TRUE \\
Begin & 1 \\
End & max\_epochs \\
\midrule
\multicolumn{2}{l}{\textbf{Data Preprocessor}} \\
Mean & [123.675, 116.28, 103.53] \\
Standard Deviation & [58.395, 57.12, 57.375] \\
BGR to RGB & TRUE \\
Padding Value & 0 \\
Segmentation Padding Value & 255 \\
Size & -10,241,024 \\
\midrule
\multicolumn{2}{l}{\textbf{Trainer}} \\
Number of Things Classes & 2 \\
Train Batch Size per GPU & 1 \\
Train Number of Workers & 4 \\
Test Batch Size per GPU & 1 \\
Test Number of Workers & 4 \\
Epochs & 100 \\
\bottomrule
\end{tabular}
}
\end{table}

\subsection{Supplementary Experiments}

\textbf{Experimental Setups.} To improve reproducibility, we present the used parameters details in Table \ref{supp-detail}.

\begin{figure}[h]
    \centering
    % First row
    \begin{subfigure}{0.3\columnwidth}
        \centering
        \includegraphics[width=\linewidth]{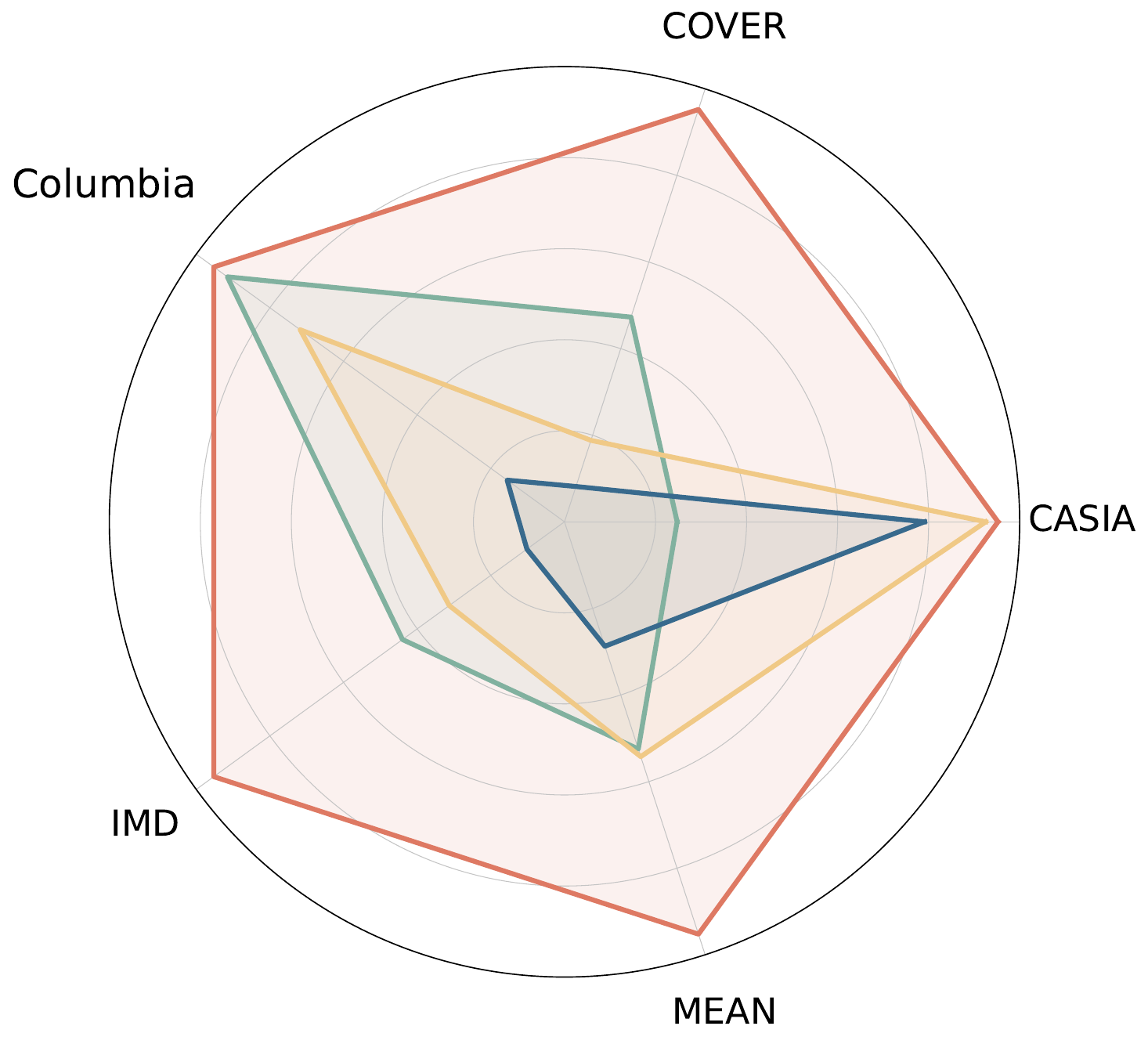}
        \caption{ I-F1}
        \label{fig:sub1}
    \end{subfigure}
    \hfill  
    \begin{subfigure}{0.3\columnwidth}
        \centering
        \includegraphics[width=\linewidth]{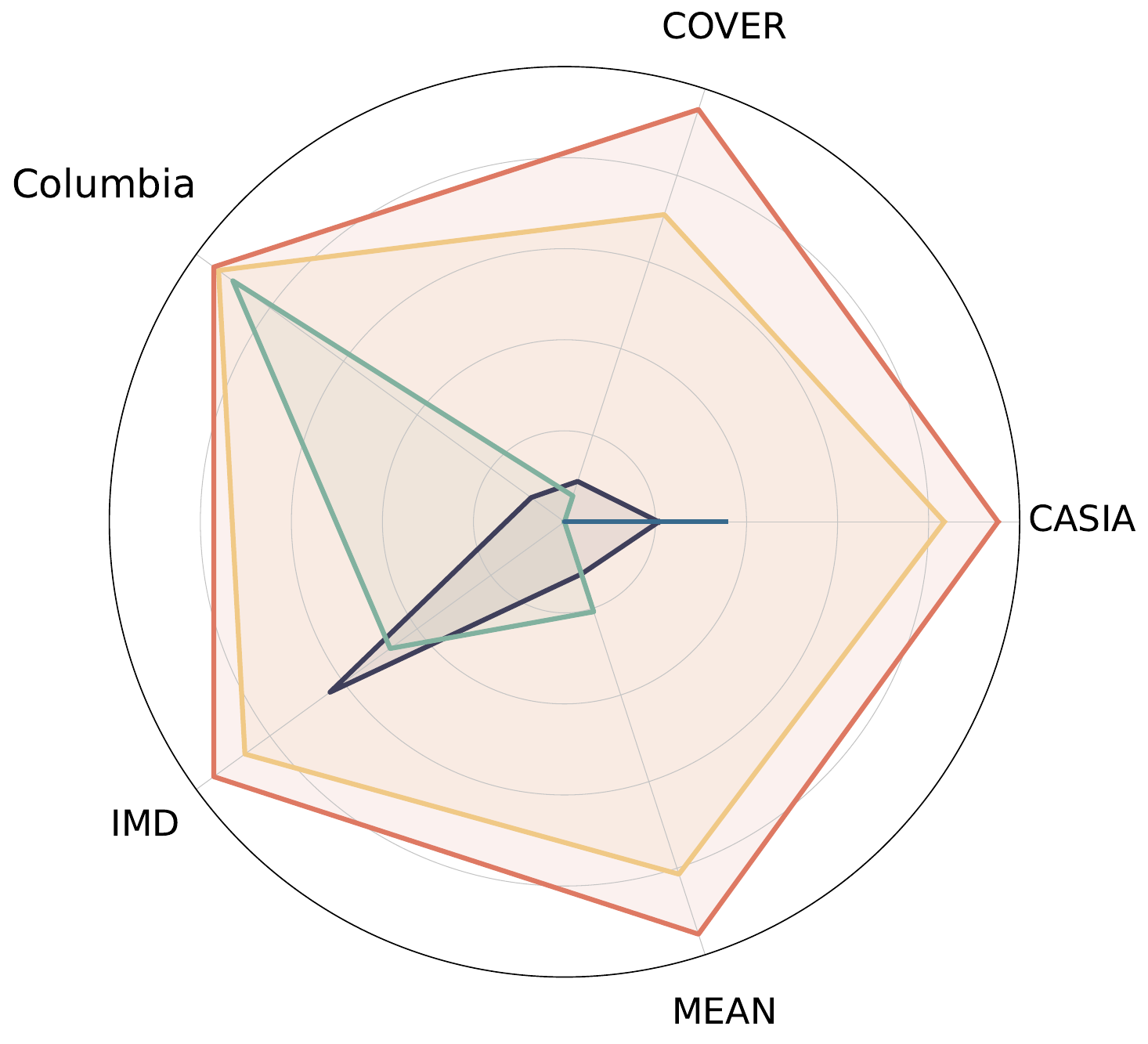}
        \caption{I-AUC}
        \label{fig:sub3}
    \end{subfigure}
    \hfill
    \begin{subfigure}{0.3\columnwidth}
        \centering
        \includegraphics[width=\linewidth]{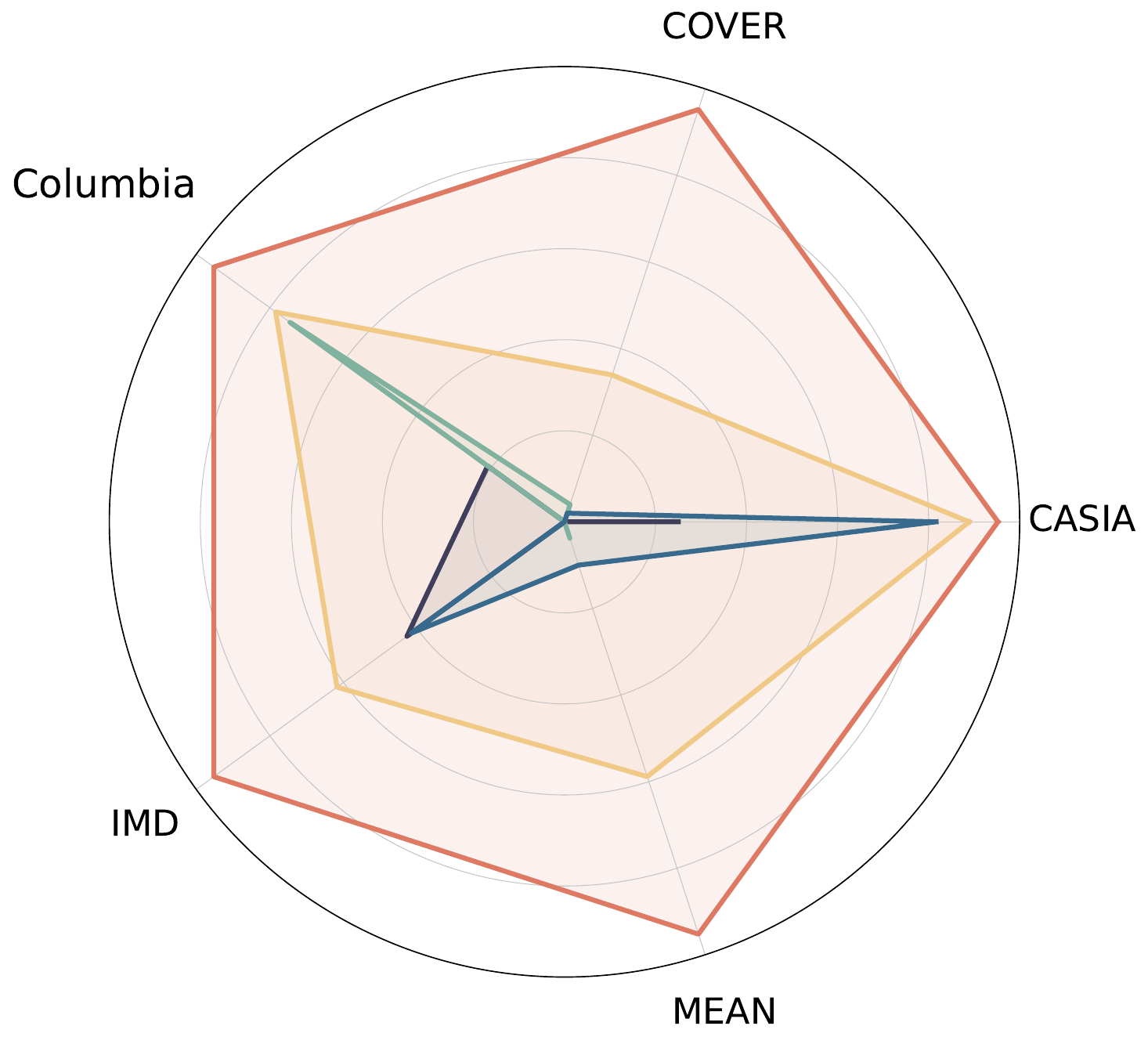}
        \caption{C-F1 }
        \label{fig:sub4}
    \end{subfigure}

    % Third row
    \begin{subfigure}{0.3\columnwidth}
        \centering
        \includegraphics[width=\linewidth]{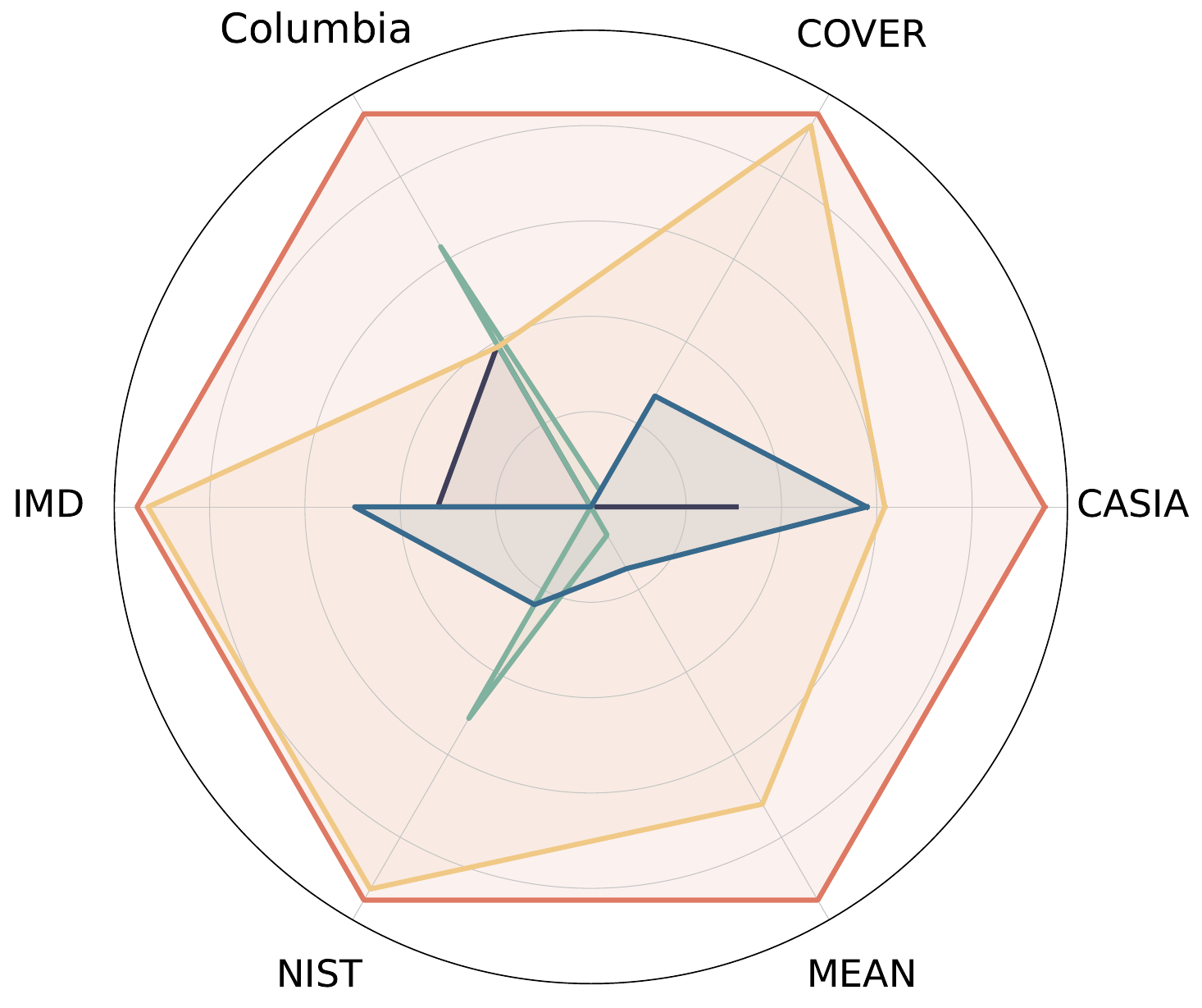}
        \caption{Best threshold P-F1 }
        \label{fig:sub5}
    \end{subfigure}
    \hfill
    \begin{subfigure}{0.3\columnwidth}
        \centering
        \includegraphics[width=\linewidth]{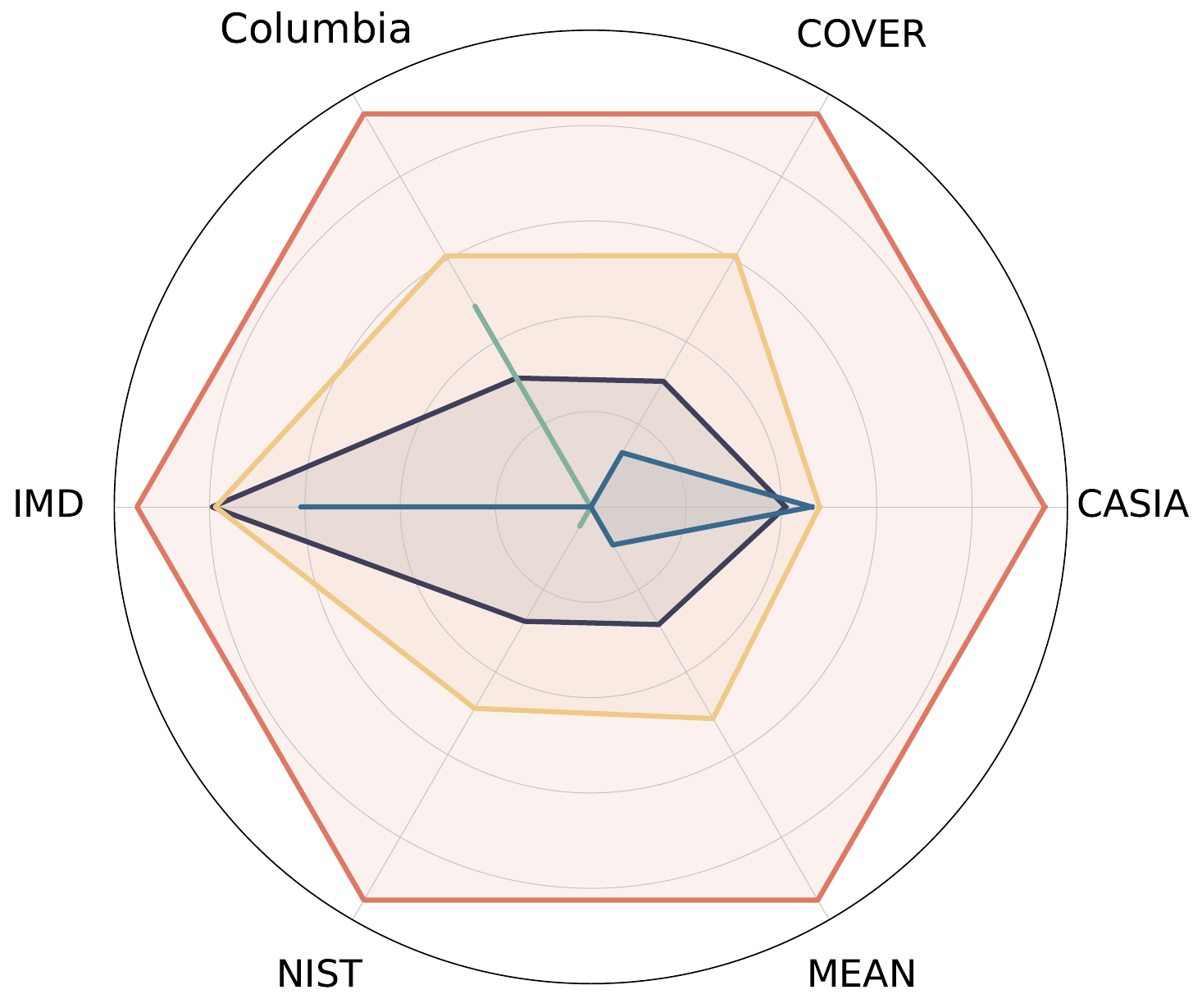}
        \caption{Fixed threshold P-F1 }
        \label{fig:sub6}
    \end{subfigure}
    \hfill
    \begin{subfigure}{0.3\columnwidth}
        \centering
        \includegraphics[width=0.6\linewidth]{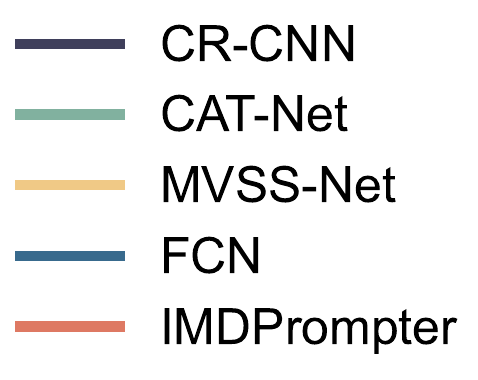}
        \caption{ }
        \label{fig:sub2}
    \end{subfigure}
    
    \caption{Visualization of Metric Data Comparison between IMDPrompter and State-of-the-Art (SOTA) Methods}
    \label{fig:grid}
\end{figure}

\textbf{Overall Performance in Detection and Localization.} Table \ref{tab:overall-detection-localization} and Figure \ref{fig:grid} shows the overall performance of pixel-level and image-level manipulation detection. We use the harmonic mean of image-level detection F1 and pixel-level localization F1, denoted as C-F1, as our overall performance metric. As shown in Table 3, IMDPrompter achieves the best performance in all settings, with a performance improvement of 45.3\% over the next best method. Notably, in the experiments on the COVER dataset, we achieved a performance gain of 110.7\%.

\begin{table*}[ht]
\centering
\caption{Overall Detection and Localization Performance (Com-F1 Score). The best results for each test set are highlighted in bold and second-best values are underlined.}
\renewcommand\arraystretch{1.2}
\resizebox{0.7\linewidth}{!}{

\begin{tabular}{lccccc}
\hline
\textbf{Method}                                       & \textbf{CASIA}               & \textbf{COVER}               & \textbf{Columbia}            & \textbf{IMD}                 & \textbf{MEAN}                 \\ \hline
H-LSTM \cite{bappy2019hybrid}        & 0.0                          & 0.0                          & 3.8                          & 0.0                          & 0.94                          \\
ManTra-Net \cite{wu2019mantra}       & 0.0                          & 0.0                          & 0.0                          & 0.0                          & 0.00                          \\
CR-CNN \cite{yang2020constrained}    & 30.3                         & 18.1                         & 41.3                         & 23.7                         & 28.35                         \\
GSR-Net \cite{zhou2018generate}      & 0.0                          & 0.0                          & 4.2                          & 0.0                          & 1.06                          \\
SPAN \cite{hu2020span}               & 0.0                          & 0.0                          & 0.0                          & 0.0                          & 0.00                          \\
CAT-Net \cite{kwon2021cat}           & 20.0                         & 19.7                         & 69.1           & 9.5                          & 29.59                         \\
MVSS-Net \cite{chen2021image}        & \underline{56.6}            & \underline{31.7}            & \underline{71.1}              & \underline{30.0}            & \underline{47.36}            \\
\rowcolor[HTML]{E8E6E6} 
FCN \cite{long2015fully}             & 53.6              & 18.9                         & 30.5                         & 23.3                         & 31.57                         \\
\rowcolor[HTML]{E8E6E6} 
\cellcolor[HTML]{E8E6E6}                              & \textbf{76.8}               & \textbf{66.8}               & \textbf{90.3}            & \textbf{41.3}               & \textbf{68.82}               \\
\rowcolor[HTML]{E8E6E6} 
\multirow{-2}{*}{\cellcolor[HTML]{E8E6E6}IMDPrompter} & {\color[HTML]{FE0000} +23.2} & {\color[HTML]{FE0000} +47.9} & {\color[HTML]{FE0000} +59.8} & {\color[HTML]{FE0000} +18.0} & {\color[HTML]{FE0000} +37.25} \\ \hline
\end{tabular}

}
\label{tab:overall-detection-localization}
\end{table*}

% \subsubsection{Robustness Assessment}

% Results are shown in Table \ref{rub-1}, \ref{rub-2} and \ref{rub-3}. The line graph for the robustness analysis is shown in Figure \ref{sensitivitybar}.

% \begin{table}[]
% \centering

% \caption{Impact of JPEG Compression and Gaussian Blur on Pixel-level F1}
% \label{rub-1}
% \begin{tabular}{@{}l|cccccc|cccccc@{}}
% \toprule
% \multirow{2}{*}{Method} & \multicolumn{6}{c|}{JPEG Compression} & \multicolumn{6}{c}{Gaussian Blurs} \\ 
% \cmidrule(l){2-13} 
%                         & 100  & 90   & 80   & 70   & 60   & 50   & 0    & 5    & 11   & 17   & 23   & 29   \\ 
% \midrule
% H-LSTM                  & 15.4 & 15.0 & 14.8 & 14.7 & 14.0 & 13.9 & 15.4 & 13.4 & 12.1 & 10.8 & 10.6 & 10.3 \\
% ManTra-Net              & 15.5 & 11.9 & 8.2  & 8.0  & 8.7  & 8.6  & 15.5 & 12.0 & 11.3 & 12.0 & 11.6 & 11.2 \\
% CR-CNN                  & 40.4 & 33.6 & 22.4 & 17.1 & 11.7 & 9.2  & 40.4 & 23.3 & 22.1 & 21.1 & 20.7 & 19.9 \\
% GSR-Net                 & 37.4 & 35.9 & 32.5 & 32.4 & 30.8 & 27.0 & 37.4 & 30.5 & 26.5 & 23.5 & 20.8 & 19.0 \\
% SPAN                    & 18.2 & 17.1 & 17.1 & 16.6 & 15.7 & 14.9 & 18.2 & 10.7 & 10.6 & 10.0 & 9.6  & 9.6  \\
% CAT-Net                 & 13.5 & 2.2  & 2.0  & 1.3  & 0.9  & 0.9  & 13.5 & 9.4  & 9.3  & 6.9  & 6.1  & 6.0  \\
% MVSS-Net                & 45.1 & 43.2 & 42.5 & 41.2 & 40.2 & 39.5 & 45.1 & 37.4 & 36.0 & 33.0 & 21.1 & 17.3 \\
% IMDPrompter             & 50.6 & 47.9 & 46.3 & 45.9 & 45.0 & 44.8 & 50.6 & 46.1 & 45.4 & 43.1 & 30.9 & 25.9 \\ 
% \bottomrule
% \end{tabular}
% \end{table}

\begin{table}[h]
\centering
\caption{Impact of JPEG Compression and Gaussian Blur on Pixel-level F1}
\label{rub-1}
\resizebox{\linewidth}{!}{
\begin{tabular}{@{}l|cccccc|cccccc@{}}
\toprule
\multirow{2}{*}{Method} & \multicolumn{6}{c|}{JPEG Compression} & \multicolumn{6}{c}{Gaussian Blurs} \\ 
\cmidrule(l){2-13} 
                        & 100  & 90   & 80   & 70   & 60   & 50   & 0    & 5    & 11   & 17   & 23   & 29   \\ 
\midrule
H-LSTM \cite{bappy2019hybrid} & 15.4 & 15.0 & 14.8 & 14.7 & 14.0 & 13.9 & 15.4 & 13.4 & 12.1 & 10.8 & 10.6 & 10.3 \\
ManTra-Net \cite{wu2019mantra}   & 15.5 & 11.9 & 8.2  & 8.0  & 8.7  & 8.6  & 15.5 & 12.0 & 11.3 & 12.0 & 11.6 & 11.2 \\
CR-CNN \cite{yang2020constrained} & 40.4 & 33.6 & 22.4 & 17.1 & 11.7 & 9.2  & 40.4 & 23.3 & 22.1 & 21.1 & 20.7 & 19.9 \\
GSR-Net \cite{zhou2018generate}   & 37.4 & 35.9 & 32.5 & 32.4 & 30.8 & 27.0 & 37.4 & 30.5 & 26.5 & 23.5 & 20.8 & 19.0 \\
SPAN \cite{hu2020span} & 18.2 & 17.1 & 17.1 & 16.6 & 15.7 & 14.9 & 18.2 & 10.7 & 10.6 & 10.0 & 9.6  & 9.6  \\
CAT-Net \cite{kwon2021cat}  & 13.5 & 2.2  & 2.0  & 1.3  & 0.9  & 0.9  & 13.5 & 9.4  & 9.3  & 6.9  & 6.1  & 6.0  \\
MVSS-Net \cite{chen2021image}   & 45.1 & 43.2 & 42.5 & 41.2 & 40.2 & 39.5 & 45.1 & 37.4 & 36.0 & 33.0 & 21.1 & 17.3 \\
IMDPrompter & 76.3 & 73.6 & 72.0 & 71.6 & 70.7 & 70.5 & 76.3 & 71.8 & 71.1 & 68.8 & 56.6 & 51.6 \\ 
\bottomrule
\end{tabular}
}
\end{table}

\begin{table}[h]
\centering
\caption{Impact of JPEG Compression and Gaussian Blur on Image-level F1}
\label{rub-2}
\resizebox{\linewidth}{!}{
\begin{tabular}{@{}l|cccccc|cccccc@{}}
\toprule
\multirow{2}{*}{Method} & \multicolumn{6}{c|}{JPEG Compression}   & \multicolumn{6}{c}{Gaussian Blur}      \\ 
\cmidrule(lr){2-7} \cmidrule(l){8-13}
                        & 100  & 90   & 80   & 70   & 60   & 50   & 0    & 5    & 11   & 17   & 23   & 29   \\ 
\midrule
H-LSTM \cite{bappy2019hybrid}  & 0.0 & 0.0 & 0.0 & 0.0 & 0.0 & 0.0 & 0.0 & 0.0 & 0.0 & 0.0 & 0.0 & 0.0 \\
ManTra-Net \cite{wu2019mantra}      & 0.0 & 0.0 & 0.0 & 0.0 & 0.0 & 0.0 & 0.0 & 0.0 & 0.0 & 0.0 & 0.0 & 0.0 \\
CR-CNN  \cite{yang2020constrained} & 24.3 & 25.8 & 24.0 & 22.4 & 20.0 & 18.4 & 24.3 & 14.3 & 13.9 & 13.7 & 12.4 & 11.4 \\
GSR-Net \cite{zhou2018generate}         & 0.0 & 0.0 & 0.0 & 0.0 & 0.0 & 0.0 & 0.0 & 0.0 & 0.0 & 0.0 & 0.0 & 0.0 \\
SPAN \cite{hu2020span} & 0.0 & 0.0 & 0.0 & 0.0 & 0.0 & 0.0 & 0.0 & 0.0 & 0.0 & 0.0 & 0.0 & 0.0 \\
CAT-Net \cite{kwon2021cat}      & 38.0 & 17.9 & 17.2 & 15.0 & 12.4 & 11.0 & 38.0 & 11.9 & 11.0 & 10.5 & 10.3 & 10.7 \\
MVSS-Net \cite{chen2021image} & 75.9 & 76.2 & 75.7 & 75.3 & 74.8 & 73.0 & 75.9 & 73.8 & 72.8 & 70.7 & 65.8 & 63.4 \\
IMDPrompter & 77.3 & 77.8 & 77.2 & 76.8 & 76.1 & 74.6 & 77.3 & 75.9 & 74.6 & 73.1 & 69.3 & 66.6 \\ 
\bottomrule
\end{tabular}
}
\end{table}

\begin{table}[h]
\centering
\caption{Impact of JPEG Compression and Gaussian Blur on Com-F1}
\label{rub-3}
\resizebox{\linewidth}{!}{
\begin{tabular}{@{}l|cccccc|cccccc@{}}
\toprule
\multirow{2}{*}{Method} & \multicolumn{6}{c|}{JPEG Compression}   & \multicolumn{6}{c}{Gaussian Blur}      \\ 
\cmidrule(lr){2-7} \cmidrule(l){8-13}
                        & 100  & 90   & 80   & 70   & 60   & 50   & 0    & 5    & 11   & 17   & 23   & 29   \\ 
\midrule
H-LSTM \cite{bappy2019hybrid} & 0.0 & 0.0 & 0.0 & 0.0 & 0.0 & 0.0 & 0.0 & 0.0 & 0.0 & 0.0 & 0.0 & 0.0 \\
ManTra-Net \cite{wu2019mantra} & 0.0 & 0.0 & 0.0 & 0.0 & 0.0 & 0.0 & 0.0 & 0.0 & 0.0 & 0.0 & 0.0 & 0.0 \\
CR-CNN \cite{yang2020constrained} & 30.3 & 29.2 & 23.2 & 19.4 & 14.8 & 12.3 & 30.3 & 17.7 & 17.0 & 16.6 & 15.5 & 14.5 \\
GSR-Net \cite{zhou2018generate} & 0.0 & 0.0 & 0.0 & 0.0 & 0.0 & 0.0 & 0.0 & 0.0 & 0.0 & 0.0 & 0.0 & 0.0 \\
SPAN \cite{hu2020span} & 0.0 & 0.0 & 0.0 & 0.0 & 0.0 & 0.0 & 0.0 & 0.0 & 0.0 & 0.0 & 0.0 & 0.0 \\
CAT-Net \cite{kwon2021cat} & 19.9 & 3.9 & 3.6 & 2.3 & 1.7 & 1.7 & 19.9 & 10.5 & 10.1 & 8.3 & 7.7 & 7.7 \\
MVSS-Net \cite{chen2021image} & 56.6 & 55.1 & 54.4 & 53.2 & 52.3 & 51.2 & 56.6 & 49.7 & 48.2 & 45.0 & 32.0 & 27.2 \\
IMDPrompter & 76.8 & 75.6 & 74.5 & 74.1 & 73.3 & 72.5& 76.8 & 73.8 & 72.8 & 70.9 &62.3 & 58.1\\ 
\bottomrule
\end{tabular}
}
\end{table}

\begin{table}[t!]
\caption{Ablation Study of the Segmenter}
\label{tab:my-table-2}
\centering
\resizebox{\textwidth}{!}{
\begin{tabular}{@{}ccccccccccccccccccc@{}}
\toprule
\multicolumn{1}{l}{} &
  \multirow{2}{*}{\textbf{Params/M}} &
  \multirow{2}{*}{\textbf{GFLOPs}} &
  \multicolumn{4}{c}{\textbf{CASIA}} &
  \multicolumn{4}{c}{\textbf{COVER}} &
  \multicolumn{4}{c}{\textbf{Columbia}} &
  \multicolumn{4}{c}{\textbf{IMD}} \\
  \cmidrule(lr){4-7} \cmidrule(lr){8-11} \cmidrule(lr){12-15} \cmidrule(lr){16-19}
 &
   &
   &
  \textbf{I-AUC} &
  \textbf{I-F1} &
  \textbf{P-F1} &
  \textbf{C-F1} &
  \textbf{I-AUC} &
  \textbf{I-F1} &
  \textbf{P-F1} &
  \textbf{C-F1} &
  \textbf{I-AUC} &
  \textbf{I-F1} &
  \textbf{P-F1} &
  \textbf{C-F1} &
  \textbf{I-AUC} &
  \textbf{I-F1} &
  \textbf{P-F1} &
  \textbf{C-F1} \\
  \midrule
\textbf{PSPNet\cite{zhao2017pyramid}} &
  365.3 &
  1610.3 &
  0.979 &
  77.4 &
  76.5 &
  76.9 &
  0.794 &
  70.3 &
  63.7 &
  66.8 &
  0.984 &
  93.7 &
  87.8 &
  90.7 &
  0.673 &
  64.1 &
  30.6 &
  41.4 \\
\textbf{SETR\cite{zheng2021rethinking}} &
  1591.6 &
  2131.6 &
  0.985 &
  78.1 &
  77.1 &
  77.6 &
  0.801 &
  70.9 &
  64.3 &
  67.4 &
  0.988 &
  94.1 &
  88.2 &
  91.1 &
  0.679 &
  64.5 &
  31.1 &
  42.0 \\
\textbf{Segformer\cite{xie2021segformer}} &
  536.3 &
  1769.3 &
  0.988 &
  78.9 &
  77.8 &
  78.3 &
  0.811 &
  71.2 &
  65.1 &
  68.0 &
  0.991 &
  94.9 &
  88.7 &
  91.7 &
  0.684 &
  65.1 &
  32.1 &
  43.0 \\
\textbf{FCN\cite{long2015fully}} &
  347.6 &
  1533.2 &
  0.978 &
  77.4 &
  76.4 &
  76.9 &
  0.795 &
  70.2 &
  63.7 &
  66.8 &
  0.984 &
  93.6 &
  87.6 &
  90.5 &
  0.670 &
  63.8 &
  30.6 &
  41.4 \\
 \bottomrule
\end{tabular}
}
\end{table}
% Please add the following required packages to your document preamble:
% \usepackage{booktabs}
% \usepackage{multirow}
% \usepackage[normalem]{ulem}
% \useunder{\uline}{\ul}{}
% \begin{landscape}
% \end{lands}
\begin{table}[t!]
\caption{Computational Complexity Overhead}
\label{tab:my-table-3}
\centering
\resizebox{\textwidth}{!}{
\begin{tabular}{@{}ccccccccccccccccccc@{}}
\toprule
\multicolumn{1}{l}{} &
  \multirow{2}{*}{\textbf{Params/M}} &
  \multirow{2}{*}{\textbf{GFLOPs}} &
  \multicolumn{4}{c}{\textbf{CASIA}} &
  \multicolumn{4}{c}{\textbf{COVER}} &
  \multicolumn{4}{c}{\textbf{Columbia}} &
  \multicolumn{4}{c}{\textbf{IMD}} \\
  \cmidrule(lr){4-7} \cmidrule(lr){8-11} \cmidrule(lr){12-15} \cmidrule(lr){16-19}
 &
   &
   &
  \textbf{I-AUC} &
  \textbf{I-F1} &
  \textbf{P-F1} &
  \textbf{C-F1} &
  \textbf{I-AUC} &
  \textbf{I-F1} &
  \textbf{P-F1} &
  \textbf{C-F1} &
  \textbf{I-AUC} &
  \textbf{I-F1} &
  \textbf{P-F1} &
  \textbf{C-F1} &
  \textbf{I-AUC} &
  \textbf{I-F1} &
  \textbf{P-F1} &
  \textbf{C-F1} \\
  \midrule
\textbf{ManTra-Net\cite{wu2019mantra}} &
  4.0 &
  1009.7 &
  0.500 &
  0.0 &
  15.5 &
  0.0 &
  0.500 &
  0.0 &
  28.6 &
  0.0 &
  0.701 &
  0.0 &
  36.4 &
  0.0 &
  0.500 &
  0.0 &
  18.7 &
  0.0 \\
\textbf{SPAN\cite{hu2020span}} &
  15.4 &
  30.9 &
  0.500 &
  0.0 &
  18.4 &
  0.0 &
  0.500 &
  0.0 &
  17.2 &
  0.0 &
  0.500 &
  0.0 &
  48.7 &
  0.0 &
  0.500 &
  0.0 &
  17.0 &
  0.0 \\
\textbf{MVSS-Net\cite{chen2021image}} &
  146.9 &
  160.0 &
  0.937 &
  75.8 &
  45.2 &
  56.6 &
  0.731 &
  24.4 &
  45.3 &
  31.7 &
  0.980 &
  80.2 &
  63.8 &
  71.1 &
  0.656 &
  35.5 &
  26.0 &
  30.0 \\
\textbf{Trufor\cite{guillaro2023trufor}} &
  67.8 &
  90.1 &
  0.916 &
  - &
  44.1 &
  - &
  0.770 &
  - &
  19.9 &
  - &
  \textbf{0.996} &
  - &
  22.3 &
  - &
  - &
  - &
  21.0 &
  - \\
\textbf{IMDPrompter} &
  347.6 &
  1533.2 &
  \textbf{0.978} &
  \textbf{77.3} &
  \textbf{76.3} &
  \textbf{76.8} &
  \textbf{0.796} &
  \textbf{70.3} &
  \textbf{63.6} &
  \textbf{66.8} &
  {0.983} &
  \textbf{93.6} &
  \textbf{87.3} &
  \textbf{90.3} &
  \textbf{0.671} &
  \textbf{63.7} &
  \textbf{30.6} &
  \textbf{41.3} \\
\textbf{IMDPrompter*} &
  85.8 &
  151.3 &
  {0.951} &
  {76.1} &
  {70.3} &
  {73.1} &
  {0.779} &
  {64.5} &
  {57.7} &
  {60.9} &
  0.980 &
  {88.3} &
  {81.6} &
  {84.8} &
  {0.667} &
  {56.9} &
  {26.9} &
  {36.5}\\
  \bottomrule
\end{tabular}
}
\label{sup_1_3}
\end{table}
% \end{landscape}
% Please add the following required packages to your document preamble:
% \usepackage{booktabs}
% \usepackage[normalem]{ulem}
% \useunder{\uline}{\ul}{}
\begin{table}[t!]
\caption{Comparison with SAM-based methods.}
\label{tab:my-table-2}
\centering
\resizebox{\textwidth}{!}{
\begin{tabular}{@{}lcccccccccccccccc@{}}
\toprule
& \multicolumn{4}{c}{\textbf{CASIA}} & \multicolumn{4}{c}{\textbf{COVER}} & \multicolumn{4}{c}{\textbf{Columbia}} & \multicolumn{4}{c}{\textbf{IMD}} \\
\cmidrule(lr){2-5} \cmidrule(lr){6-9} \cmidrule(lr){10-13} \cmidrule(lr){14-17}
& \textbf{I-AUC} & \textbf{I-F1} & \textbf{P-F1} & \textbf{C-F1} & \textbf{I-AUC} & \textbf{I-F1} & \textbf{P-F1} & \textbf{C-F1} & \textbf{I-AUC} & \textbf{I-F1} & \textbf{P-F1} & \textbf{C-F1} & \textbf{I-AUC} & \textbf{I-F1} & \textbf{P-F1} & \textbf{C-F1} \\
\midrule
\textbf{MedSAM\cite{ma2024segment}} & 0.861 & 70.3 & 56.4 & 62.6 & 0.563 & 26.3 & 22.1 & 24.0 & 0.791 & 50.1 & 24.6 & 33.0 & 0.513 & 26.1 & 22.4 & 24.1 \\
\textbf{MedSAM-Adapter\cite{wu2023medical}} & 0.877 & 71.0 & 57.3 & 63.4 & 0.543 & 21.6 & 20.3 & 20.9 & 0.742 & 46.6 & 22.1 & 30.0 & 0.494 & 25.1 & 21.6 & 23.2 \\
\textbf{AutoSAM\cite{shaharabany2023autosam}} & 0.843 & 69.1 & 55.1 & 61.3 & 0.513 & 18.9 & 18.6 & 18.7 & 0.772 & 49.1 & 28.4 & 36.0 & 0.523 & 27.9 & 22.9 & 25.2 \\
\textbf{SAMed\cite{zhang2023customized}} & 0.843 & 69.9 & 52.3 & 59.8 & 0.571 & 27.6 & 24.6 & 26.0 & 0.801 & 53.1 & 31.7 & 39.7 & 0.502 & 27.6 & 23.1 & 25.2 \\
\textbf{IMDPrompter} & \textbf{0.978} & \textbf{77.3} & \textbf{76.3} & \textbf{76.8} & \textbf{0.796} & \textbf{70.3} & \textbf{63.6} & \textbf{66.8} & {0.983} & \textbf{93.6} & \textbf{87.3} & \textbf{90.3} & \textbf{0.671} & \textbf{63.7} & \textbf{30.6} & \textbf{41.3} \\
\textbf{IMDPrompter*} & {0.951} & {76.1} & {70.3} & {73.1} & {0.779} & {64.5} & {57.7} & {60.9} & 0.980 & {88.3} & {81.6} & {84.8} & {0.667} & {56.9} & {26.9} & {36.5} \\
\bottomrule
\end{tabular}
}
\end{table}

% Please add the following required packages to your document preamble:
% \usepackage{booktabs}
\begin{table}[t!]
\caption{The proportion of each view selected as the optimal view}
\label{tab:my-table-3}
\centering
\begin{tabular}{@{}lcccc@{}}
\toprule
\textbf{Prompt View} & \textbf{RGB} & \textbf{SRM} & \textbf{Bayer} & \textbf{Noiseprint} \\
\midrule
\textbf{Percent} & 13.9\% & 18.4\% & 21.4\% & 46.3\% \\
\bottomrule
\end{tabular}
\end{table}

% Please add the following required packages to your document preamble:
% \usepackage{booktabs}
% \usepackage{multirow}
\begin{table}[t!]
\caption{Performance of Generative Image Manipulation Detection}
\label{tab:my-table-4}
\centering
\renewcommand{\arraystretch}{1}
\resizebox{0.75\textwidth}{!}{
\begin{tabular}{@{}cccccc@{}}
\toprule
\multirow{2}{*}{\textbf{Method}} & \multicolumn{4}{c}{\textbf{CocoGlide}} \\
\cmidrule(lr){2-5}
                                 & \textbf{P-F1(best)} & \textbf{P-F1(fixed)} & \textbf{I-AUC} & \textbf{I-Acc}  \\ \midrule
\textbf{ManTraNet\cite{wu2019mantra}}               & 0.673               & 0.516                & 0.778          & 0.500      \\
\textbf{SPAN\cite{hu2020span}}                    & 0.35                & 0.298                & 0.475          & 0.491       \\
\textbf{MVSS-Net\cite{chen2021image}}                & 0.642               & 0.486                & 0.654          & 0.536       \\
\textbf{Trufor\cite{guillaro2023trufor}}                  & 0.72                & 0.523                & 0.752          & 0.639        \\
\textbf{IMDPrompter}             & 0.746               & 0.539                & 0.781          & 0.652       \\
\bottomrule
\end{tabular}
}
\end{table}
\begin{figure}[t!]
	\centering
    \includegraphics[width=\textwidth, keepaspectratio]{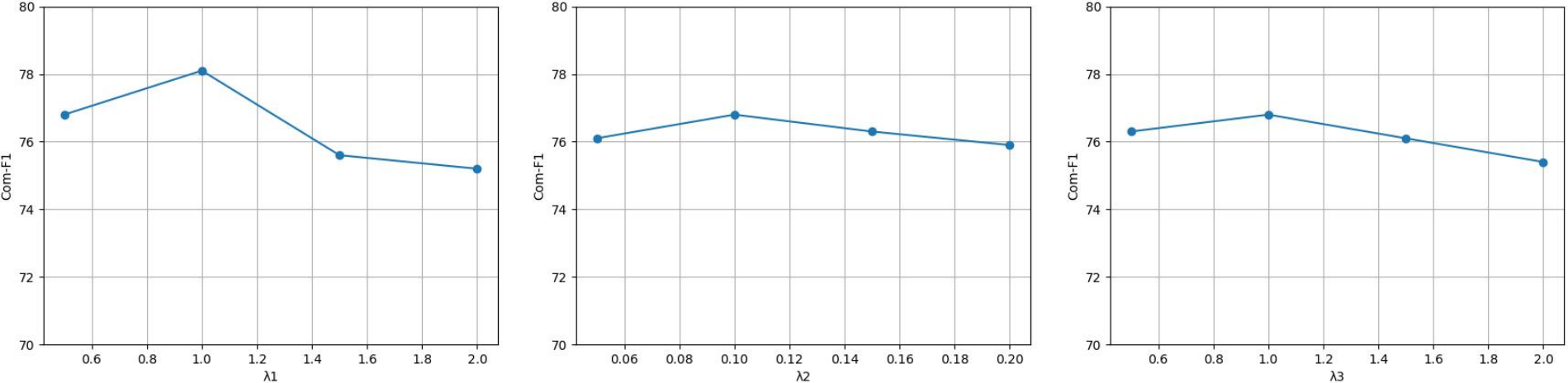}

	\caption{Hyperparametric analysis of $\lambda_1$, $\lambda_2$, and $\lambda_3$}
 
	\label{Hyperparametric}
\end{figure}

\begin{figure}[h]
	\centering
	\includegraphics[width=\linewidth]{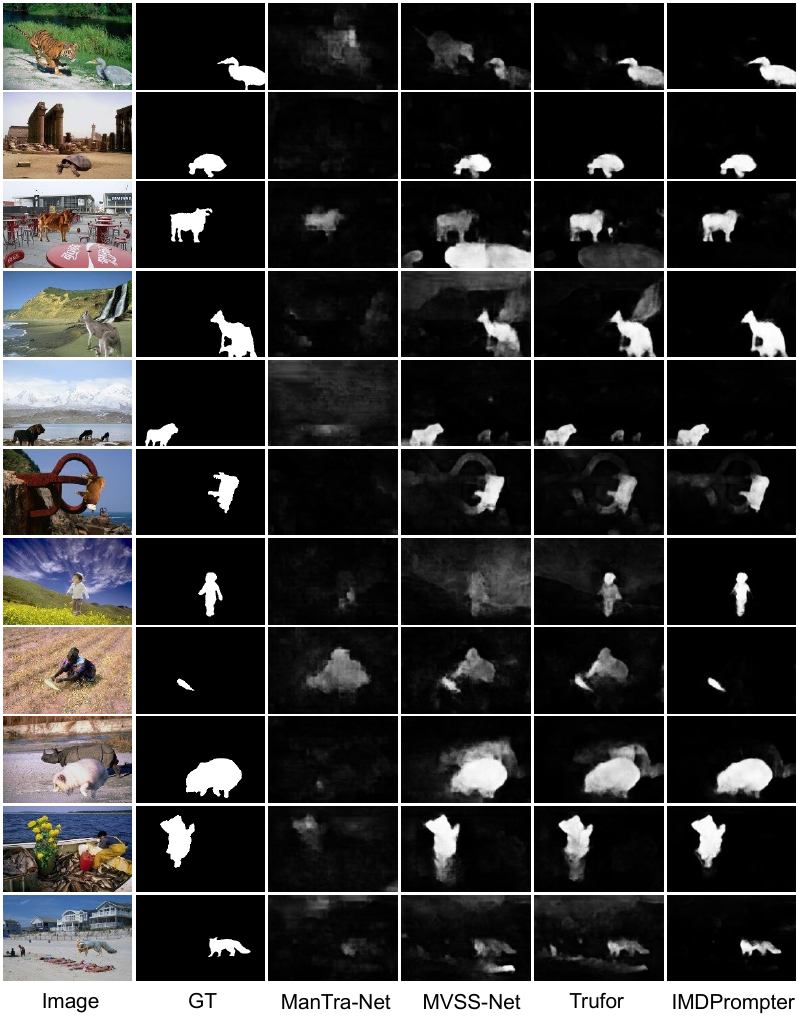}
	\caption{Some qualitative results, compared with the state-of-the-art}
	\label{vis}
\end{figure}
\begin{figure}[h]
	\centering
	\includegraphics[width=\linewidth]{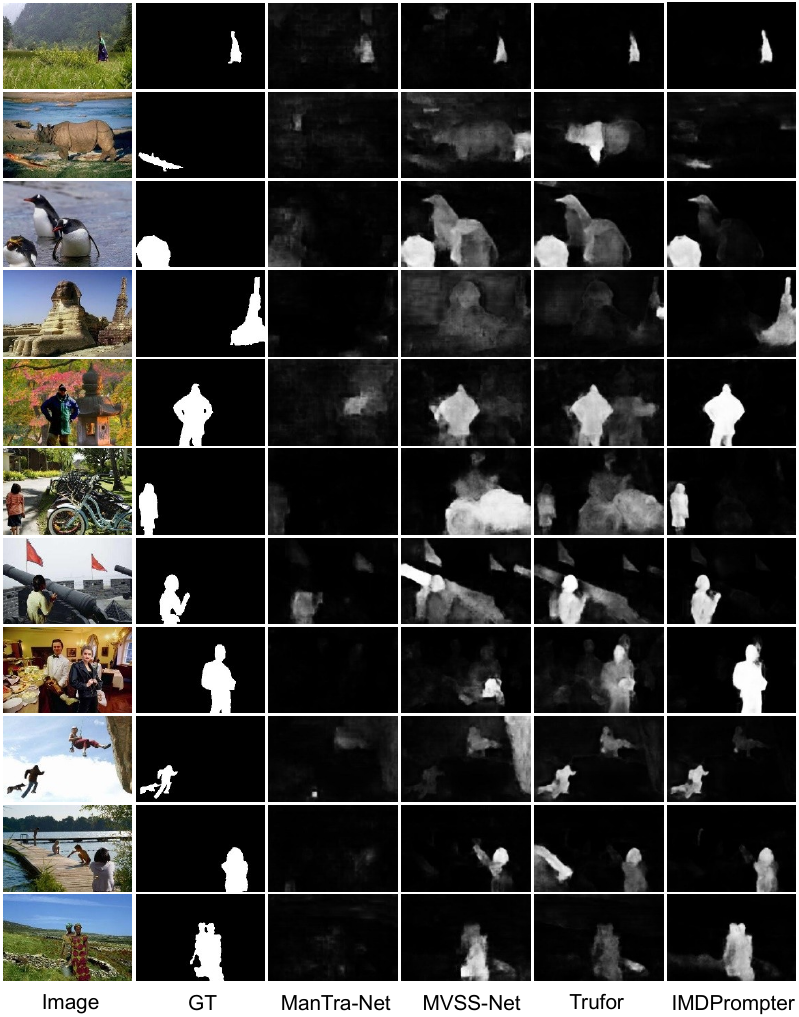}
	\caption{Some qualitative results, compared with the state-of-the-art}
	\label{vis}
\end{figure}
\begin{figure}[h]
	\centering
	\includegraphics[width=\linewidth]{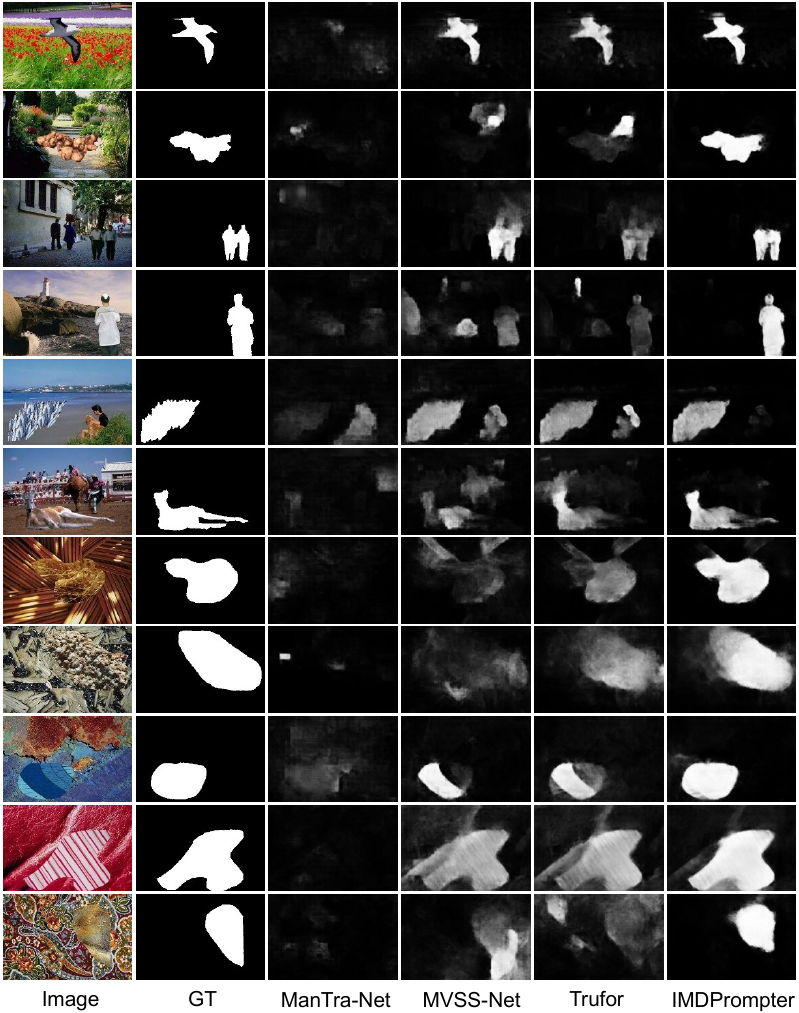}
	\caption{Some qualitative results, compared with the state-of-the-art}
	\label{vis}
\end{figure}

\textbf{Ablation study of the segmenters for the four prompt views: }We selected several segmenters, including FCN, PSPNet, SETR, and Segformer, for ablation experiments. As shown in the table below, the lightweight FCN based on MobileNet is sufficient to effectively guide SAM's automated prompt learning. The introduction of more powerful segmenters like PSPNet, SETR, and Segformer did not result in a significant improvement in detection performance but instead introduced greater computational overhead. Therefore, we chose the lightweight FCN as our segmenter.\par

\textbf{Computational Complexity Analysis:} As shown in Table \ref{sup_1_3}, I-AUC represents Image-level AUC, I-F1 represents Image-level F1, P-F1 represents Pixel-level F1, and C-F1 represents the harmonic mean of I-F1 and P-F1. It is worth noting that the IMDPrompter mentioned in the main text is based on SAM-H, which has a high computational complexity. Therefore, we implemented a lightweight version, IMDPrompter*, based on Mobile SAM. As can be seen from Table \ref{sup_1_3}, IMDPrompter achieved the best performance on almost all metrics, and IMDPrompter* achieved performance second only to IMDPrompter in nearly all metrics.

\textbf{Comparison with SAM-based Methods:} As shown in Table \ref{sup_1_3}, We supplement the performance of four SAM-based image segmentation methods: MedSAM, MedSAM-Adapter, AutoSAM, and SAMed. As we can see, these four SAM-based methods exhibit limited performance in the iimage manipulation detection task due to their reliance solely on semantically relevant information. In particular, there is a significant performance gap between them and IMDPrompter, especially on the out-of-domain test sets CVOER, Columbia, and IMD.

\textbf{Proportion of Four Views Selected as Optimal Prompts.} As shown in Table \ref{tab:my-table-3}, we summarize the proportion of each prompt view being selected as the optimal prompt during training. We observe that the RGB view is selected as the optimal prompt in only 13.9\% of cases, indicating that semantically relevant features alone are insufficient to guide SAM's prompt learning. On the other hand, the Noiseprint view is chosen as the optimal prompt in 46.3\% of cases, demonstrating that the semantically irrelevant information provided by Noiseprint is the most crucial.

\textbf{Hyperparametric Analysis} As shown in Figure \ref{Hyperparametric}, we conducted hyperparameter analysis on $\lambda_1$,$\lambda_2$, and $\lambda_3$, ultimately selecting the optimal parameter configuration: $\lambda_1=1.0, \lambda_2=0.1, \lambda_3=1.0 $.

\textbf{Performance of Generative Image Manipulation Detection:} Following the work of Trufor, we supplement the performance of IMDPrompter on CocoGlide (image tampering based on diffusion models). As shown in the table below, our IMDPrompter also achieves objective results for AIGC-edited images.

\textbf{Impact of Quality Degradation.} As shown in Table \ref{rub-1},\ref{rub-2},\ref{rub-3}, following \cite{dong2022mvss}, we evaluated the robustness of the models under two common image processing operations encountered during the dissemination of images on the internet, namely JPEG compression and Gaussian blur. Comparing these two operations, Gaussian blur has a more significant impact on detection performance, especially when using larger 17x17 scale convolution kernels. Compared to previous methods, our IMDPrompter exhibits better robustness.

\textbf{More Visualization.} As shown in Figure \ref{vis}, we have included additional visualizations.

\end{document}